# Closing Gaps in Emissions Monitoring with Climate TRACE

## Authors

Brittany V. Lancellotti[1*], Jordan M. Malof[2], Aaron Davitt[3], Gavin McCormick[3], Shelby Anderson[4], Pol Carbó-Mestre[5], Gary Collins[4], Verity Crane[6], Zoheyr Doctor[3], George Ebri[6], Kevin Foster[4], Trey M. Gowdy[1], Michael Guzzardi[6], John Heal[6], Heather Hunter[4], David Kroodsma[8], Khandekar Mahammad Galib[6], Paul J. Markakis[7], Gavin McDonald[5], Daniel P. Moore[3], Eric D. Nguyen[3], Sabina Parvu[6], Michael Pekala[4], Christine D. Piatko[4], Amy Piscopo[3], Mark Powell[9], Krsna Raniga[3], Elizabeth P. Reilly[4], Michael Robinette[4], Ishan Saraswat[3], Patrick Sicurello[4], Isabella Söldner-Rembold[6], Raymond Song[10], Charlotte Underwood[6], & Kyle Bradbury[1]

## Affiliations

[1]Nicholas Institute for Energy Environment & Sustainability, Duke University, Durham, NC, USA.
[2]Department of Electrical Engineering and Computer Science, University of Missouri, Columbia, MO, USA.
[3]WattTime, Oakland, CA, USA.
[4]Johns Hopkins Applied Physics Laboratory, Laurel, MD, USA.
[5]Marine Science Institute, University of California, Santa Barbara, Santa Barbara, CA, USA.
[6]TransitionZero, London, UK.
[7]Duke University, Durham, NC, USA.
[8]Global Fishing Watch, Washington, DC, USA.
[9]Global Fishing Watch, Bainbridge Island, WA, USA.
[10]CTrees, Pasadena, CA, USA.
*Corresponding author, Brittany.Lancellotti@duke.edu

## Abstract

Global greenhouse gas emissions estimates are essential for monitoring and mitigation planning. Existing emissions datasets provide critical foundations for understanding emissions patterns across sectors, geographies, and time scales. Through a structured assessment of recent emissions datasets, we identified opportunities to further increase the actionability of emissions data through more comprehensive source-level coverage, finer spatial and temporal resolution, and more frequent updates. Building on existing resources to address these opportunities, we present the Climate TRACE framework and resulting dataset, which is available on an open-access platform (climatetrace.org). The Climate TRACE framework synthesizes existing emissions data, prioritizing accuracy, coverage, and resolution, and fills remaining gaps using sector-specific estimation approaches. The resulting dataset is the first to provide global emissions estimates for individual sources (e.g., individual power plants) for most anthropogenic emitting sectors. The dataset spans January 1, 2021, to the present, with a two-month reporting lag and monthly updates. This dataset and open-access platform provides access to detailed emissions estimates for most subnational governments worldwide. By combining source-level spatial detail, monthly updates, and broad sectoral coverage, the dataset is designed to support analyses relevant to emissions monitoring and mitigation planning.

## 1. Introduction

Without dramatic intervention, the latest Intergovernmental Panel on Climate Change (IPCC) report predicts a global mean surface temperature increase of at least 1.5°C before the end of the century due to the release of anthropogenic greenhouse gases (GHGs). The impacts of global temperature rise, including extreme weather, sea level rise, and species extinction, are expected to persist for centuries to millennia, demanding rapid and effective action to reduce anthropogenic GHG emissions [1]. The worldwide adoption of successful mitigation policies has been projected to result in considerable emissions reductions that would help limit temperature rise to 2°C (2). However, the development of effective mitigation strategies hinges on accurate emissions tracking and the prioritization of reduction targets [3, 4, 5, 6].

Global GHG datasets address this need by tracking emissions, providing a quantitative basis for assessing the effectiveness of mitigation strategies, and promoting accountability [7, 8, 9]. For example, national emissions datasets, most prominently those reported to the United Nations Framework Convention on Climate Change (UNFCCC), are used to assess the progress of global emissions mitigation strategies [10]. GHG accounting is also embedded in carbon pricing mechanisms and regulatory frameworks worldwide [6].

Although global GHG datasets provide a critical pathway for emissions monitoring and tracking mitigation progress [4, 11, 12], most existing datasets have characteristics that limit their actionability. These include reliance on self-reported data that may be inaccurate, lack of sufficient global coverage and resolution, infrequent updates, and limited user accessibility [6, 11, 13−17]. While these limitations are widely recognized, few assessments have taken a comprehensive approach to evaluating the emissions data landscape. Most existing assessments are outdated or focused on identifying quantitative inconsistencies across a small set of datasets, frequently within a single emissions sector [13, 18−21].

### 1.1 Contributions

In this work, we present Climate TRACE, a global framework that integrates existing data when possible and provides newly generated emissions data when necessary, resulting in a single, comprehensive, high-resolution and regularly updated dataset. To motivate this work, we first provide a structured assessment of recent emissions datasets, highlighting their respective strengths in advancing emissions estimation, while also identifying opportunities to enhance actionability. These include improvements in resolution (spatial and temporal), coverage (spatial, temporal, and sectoral), and latency. Building on this assessment, we then introduce the Climate TRACE dataset, which results from combining existing datasets to leverage their complementary strengths, systematically prioritizing the highest-quality available information. Where data gaps remain, the framework leverages new or existing methodologies for generating new estimates that fill gaps in resolution, coverage, and timeliness.

## 2. The current emissions data landscape

We first present a conceptual model for comparing existing global GHG emissions datasets across characteristics that impact their actionability (Table 1). We assess the current emissions data landscape according to this conceptual model and synthesize our findings to reveal data gaps that have the potential to reduce actionability. We aim to include all existing emissions datasets with global coverage that report emissions for more than one sector, drawing on the most recent official dataset documentation and/or peer-reviewed literature available for each (see Supplementary Table 1).

**Table 1**. **Framework for evaluating emissions datasets**. Conceptual model for comparing existing global greenhouse gas emissions datasets and how those characteristics relate to potential pathways to improve their actionability. The leftmost column describes a potential pathway for driving data actionability. The two subsequent columns describe (1) the associated measurable dataset characteristic and (2) the dimensions along which we evaluated those characteristics.

| Potential pathways to more effective emissions mitigation | Measurable dataset characteristic | Dimensions of comparison for the dataset characteristic |
|---|---|---|
| Provide accurate emissions data and information on where emissions occur, both of which could be used to develop and implement more effective mitigation plans. | Accuracy | - Inclusion of direct accuracy evaluation, which includes comparison of emissions against in situ values<br>- Independence from self-reported data |
| Provide a centralized, comprehensive picture of emissions across time, space, and emissions sector, enabling coordinated action and progress tracking to meet international agreements. | Coverage or completeness | - Spatial<br>- Temporal<br>- Sectoral<br>- Pollutant |
| Enable localized climate action: provide detailed emissions data for subnational actors lacking datasets. Improve city, state, corporate, and investor action. | Resolution | - Spatial<br>- Temporal |
| Establish more effective carbon markets. | | |
| Enable timely evaluation and adjustment of current policies to more effectively meet emissions goals. | Update frequency | - Update frequency |
| Widespread climate action: open access to and broad adoption of emissions data to allow diverse decision-makers and experts to create data-driven emissions reduction strategies. | Usability | - Accessibility/inclusivity<br>- Visualization |

### 2.1 Accuracy

We define accuracy as the degree to which estimated emissions match the true emissions values. Because true emissions values are rarely directly observable, most datasets cannot directly compute accuracy for all estimates [3, 5]. We therefore compare datasets based on whether they address accuracy more directly by evaluating estimates against in situ (i.e., direct) measurements, which are expected to have low uncertainty and bias. We also indicate whether datasets integrate self-reported data, which are often error-prone [18].

Of the 20 global emissions datasets we reviewed, 15 report cross-dataset comparison as their primary means of assessing accuracy (Supplementary Table 2). Cross-dataset comparisons cannot always guarantee accuracy if the accuracy of the reference datasets themselves has not been confirmed. Yet due to the scarcity of in situ activity (i.e., processes that contribute to GHG generation or release) and emissions measurements, this approach is often the best available. This is true of widely used datasets including the Emissions Database for Global Atmospheric Research (EDGAR [22]), the Community Emissions Data System (CEDS [23]), IEA [24], and UNFCCC [25], among others. Three of the 20 datasets we reviewed did not specify a formal accuracy assessment approach (Supplementary Table 2). While it is possible that some accuracy assessment occurs outside of published documentation, the absence of reported methods and results inhibits stakeholders from evaluating the accuracy of the data.

Only a small subset of datasets, including Carbon Monitor [26], CDIAC-FF [27], and the Global Carbon Budget [28], supplement cross-dataset evaluation with a comparison of emissions components (e.g., activity, emission factors) against in situ data (Supplementary Table 2). For example, Carbon Monitor compares its ground transportation model against real traffic counts from roads in Paris and cross-checks residential emissions against measured natural gas consumption from several European countries [26]. Datasets that provide inversion-modeled estimates, like CarbonTracker [29], offer a more direct assessment of accuracy by comparing to atmospheric measurements that can have high uncertainty but do not rely on emitter self-reporting.

While the accuracy of inversion-modeled estimates can be assessed more directly, inversion models typically estimate the net balance of emissions and removals over space and time and do not provide detail on individual emissions sectors or sources.

#### 2.1.2 Incorporation of self-reported data

All of the datasets we reviewed incorporate self-reported emissions or activity data in at least some of their estimates (Supplementary Table 2). For example, the UNFCCC inventory compiles emitter-reported estimates or derives them from self-reported energy use statistics (e.g., International Energy Agency [IEA]). These self-reported sources are essential to achieving broad coverage and reporting consistency across countries and sectors. Yet self-reporting of emissions and activity is prone to inaccuracies from inconsistent accounting methods, limited accounting resources, and potential under-reporting [18]. Inaccuracies in self-reported data may be propagated throughout multiple global datasets, hindering the ability to manage emissions effectively. For example, any discrepancies in IEA data would be present in EDGAR [22] and CEDS [23].

### 2.2 Coverage and completeness

Although many datasets are described as having global coverage, closer examination often reveals gaps that complicate cross-dataset comparisons and may distort global emissions totals [18]. For example, global datasets that collect self-reported emissions through surveys have been found to omit entire nations, commonly low- and middle-income countries [30].

Time coverage also varies widely across global datasets (Table 2), impacting the usability of emissions data and can complicate cross-dataset comparisons. Shorter time spans may be too limited to identify long-term patterns, while historical datasets often rely on extrapolation methods that can be less accurate. Additionally, historical data (e.g., 1750–2000) may be less useful for decision-making today, compared to a dataset with more recent, though shorter, time coverage (e.g., 2020–2025).

**Table 2**. **Overview of existing emissions datasets**. Comparison of existing global greenhouse gas datasets across several dimensions, including only those that are easily quantifiable or exhibit significant heterogeneity across inventories.

| Global greenhouse gas emissions datasets | Spatial resolution | | | | | | | Temporal resolution | | | | Update frequency | | | | Greenhouse gases included | | | Period of time coverage | | | | Sectoral coverage | | |
|---|---|---|---|---|---|---|---|---|---|---|---|---|---|---|---|---|---|---|---|---|---|---|---|---|---|
| | Country | District | 111.1 km² grid | 55.5 km² grid | 11.1 km² grid | < 11.1 km² grid | Asset | Annual | Monthly | Daily | Hourly | 2 - 3 yr | 1.5 - 2 yr | 1 yr | 1-6 mos. | $CO_2$ | + $CH_4$, $N_2O$ | + non-GHGs | Pre-1950 | 1950-1999 | 2000-2020 | 2020 + | Fossil fuel combustion | + All other energy emissions | + All other anthropogenic emissions |
| Climate TRACE (v5.4.1) | | | | | | | | | | | | | | | | | | | | | '15 | '26 | | | |
| Stat. Review of World Energy (2025) | | | | | | | | | | | | | | | | | | | | '65 | | '24 | | | |
| CAMS-GLOB-ANT (v6.2) | | | | | | | | | | | | | | | | | | | | | '00 | '26 | | | |
| Carbon Monitor | | | | | | | | | | | | | | | | | | | | | '19 | '25 | * | | |
| CarbonTracker (CT2022) | | | | | | | | | | | | | | | | | | | | | | | | | |
| CDIAC-FF | | | | | | | | | | | | | | | | | | | 1750 | | '22 | | * | | |
| CEDS v2025.03.18 | | | | | | | | | | | | | | | | | | | 1750 | | '23 | | | | |
| Climate Watch | | | | | | | | | | | | | | | | | | | | '90 | | '23 | | | |
| EDGAR (2025) | | | | | | | | | | | | | | | | | | | | '70 | | '24 | | | |
| EIA Int'l Energy Outlook (2023)^ | | | | | | | | | | | | | | | | | | | '49 | | | '22 | | | |
| FAOSTAT | | | | | | | | | | | | | | | | | | | | '61 | | '23 | | | |
| FFDAS (v2.2) | | | | | | | | | | | | | | | | | | | | '97 | '15 | | | | |
| GCP-GridFED (v2024.0) | | | | | | | | | | | | | | | | | | | | '59 | | '23 | * | | |
| GID | | | | | | | | | | | | | | | | | | | | '70 | | '22 | * | | |
| Global Carbon Budget (v2025) | | | | | | | | | | | | | | | | | | | 1750 | | | '24 | * | | |
| GRACED | | | | | | | | | | | | | | | | | | | | | '19 | '25 | * | | |
| IEA GHG Emissions from Energy (2025) | | | | | | | | | | | | | | | | | | | | '60 | '23 | | | | |
| ODIAC2024† | | | | | | | | | | | | | | | | | | | | | '00 | '23 | * | | |
| PIK PRIMAP-hist (v2.7) | | | | | | | | | | | | | | | | | | | 1750 | | | '24 | | | |
| UNFCCC | | | | | | | | | | | | | | | | | | | | '90 | | '21 | | | |

| | | | | | |
|---|---|---|---|---|---|
| | Could be derived from original dataset | | Released as single dataset | | 10 km² – 10m² (varies by sub-sector) |
| | Represents cumulative inclusion | | Includes only agricultural emissions | | Highest provided |

†Dataset includes a few sectors beyond fossil fuel combustion and cement calcination.
*Dataset includes emissions from cement calcination in addition to fossil fuel combustion. Supplementary Table 1 defines dataset acronyms.
^Dataset is extrapolated to the year 2050.

Emissions sector coverage also differs markedly across global datasets (Table 2). A few global datasets, like EDGAR, support comprehensive emissions tracking across all sectors, but these often have coarse spatial resolution and infrequent updates, limiting their actionability. In contrast, the IEA dataset covers emissions from energy only, and users seeking comprehensive data would need to consult other sources. Datasets also have a range of GHG pollutant coverage (Table 2), impacting their usability.

### 2.3 Resolution and update frequency

Several global datasets estimate emissions with the finest spatial granularity at the country level, while others present emissions as high-resolution spatial grids (Table 2). Gridded datasets can serve as inputs to general circulation models and atmospheric chemical transport models, which project climate change and forecast air quality. Some datasets provide asset-level estimates (e.g., power plants, major manufacturing facilities) but do not have global coverage or are restricted to a few sectors. For example, the Global Infrastructure Emissions Detector (GID) dataset provides asset-level $CO_2$ emissions from the power, iron and steel, and cement sectors with global coverage for 1990–2022 [31]. Bun and others [32] provided asset-level estimates for Poland with comprehensive sectoral coverage for 2010, and the Vulcan Project Fossil Fuel $CO_2$ dataset estimates facility-level $CO_2$ emissions from fossil fuel combustion and cement production within the USA for 2010–2021, presented as a 1 $km^2$ grid [33]. While they provide detailed emissions for their specific regions, these spatially or sectorally limited datasets do not enable global comparisons or tracking.

Global datasets have variable temporal resolution and update frequencies. Several offer high-frequency estimates but have more limited sectoral coverage (e.g., Carbon Monitor, CarbonTracker). Few datasets are updated more than once a year, restricting timely emissions tracking and evaluations of mitigation progress (Table 2).

### 2.4 Usability

We evaluated usability of emissions estimates by examining each dataset's format, retrieval process, and the availability of interpretable data displays. While all the datasets we assessed are freely available to download (with the exception of IEA data [24]), they vary in accessibility and format. Some datasets require additional processing steps or are hosted on platforms that require coding skills to access, such as through application programming interfaces. For example, several datasets, like CarbonTracker [29], Fossil Fuel Data Assimilation System (FFDAS) [34], Global Carbon Project Gridded Fossil Emissions Dataset (GCP-GridFED) [35], and Open-source Data Inventory for Anthropogenic $CO_2$ (ODIAC) [36], provide emissions estimates primarily in spatially referenced formats, which require technical skills for visualization and processing. Similarly, most existing datasets offer downloadable data without an integrated, user-friendly visualization interface, while a few (Carbon Monitor [26], Climate Watch [37], FAOSTAT [38], and GID [31]) provide interactive emissions maps that facilitate interpretation for a broader user base. Expanding the usability of these datasets through open visualization and analytical features could empower a broader range of stakeholders to make informed decisions.

### 2.5 Opportunities for enhancement

Our findings indicate that existing global GHG emissions datasets have several limitations, including limited and inconsistent accuracy assessment, reliance on self-reported data, incomplete spatial or sectoral coverage, lack of comprehensive asset-level estimates, infrequent or inconsistent data updates, and limited accessibility. We argue that enhancing individual datasets along several dimensions and combining them to fill spatial, temporal, and sectoral gaps can significantly improve the actionability of emissions data, which motivates the development of Climate TRACE.

## 3. The Climate TRACE framework

We introduce Climate TRACE (Tracking Real-time Atmospheric Carbon Emissions), a framework that harmonizes the fragmented global emissions data landscape into a single, openly accessible dataset. It does so by integrating the most accurate data sources available, filling gaps with newly generated sector-specific estimates, and prioritizing accuracy, coverage, and resolution. The resulting dataset addresses a fundamental limitation of the existing data landscape: no single dataset provides the spatial, temporal, sectoral, and pollutant coverage that comprehensive emissions accounting and climate action require. This forces stakeholders to consult multiple inconsistent sources. Climate TRACE resolves this by providing the first globally comprehensive asset-level estimates across most anthropogenic emitting sectors (see Supplementary Dataset 1) from January 1, 2021, to the present, and country-level estimates from January 1, 2015, to the present. Both are updated monthly with a two-month reporting lag. Below, we assess the Climate TRACE dataset along the dimensions included in our conceptual model (Table 1). We contextualize each feature within the wider landscape of existing datasets, demonstrating how this harmonization improves interoperability, and ultimately, the actionability of emissions information.

### 3.1 Accuracy

Accuracy remains a major challenge across global emissions datasets due to the limited availability of in situ measurements. Yet stakeholders have identified the accuracy of emissions estimates as essential for enabling data-driven action [39]. To address this challenge, Climate TRACE leverages the strengths of existing datasets, prioritizing integration of the most accurate emissions data using a structured framework (see section 5). Broadly, the framework evaluates accuracy based on direct comparisons to in situ data. When this is not possible, the framework employs proxy accuracy evaluation approaches. These include (1) comparing a component of the emissions estimation process (e.g., activity) against in situ data and (2) comparing final emissions estimates to those of other datasets to contextualize accuracy. Although we try to minimize such occurrences, in some rare cases, neither in situ measurements nor viable proxies for accuracy assessment are available. These accuracy assessment approaches operate at both the input and output stages of estimation, increasing the likelihood that integrated estimates are more accurate than any individual dataset.

Climate TRACE's accuracy assessment procedures vary considerably by emissions sub-sector and are documented in full in Supplementary Dataset 2. We note that, as of May 2026, approximately 81% of Climate TRACE's estimated total global emissions are derived from peer-reviewed methodologies (Table 3), and we are continually working toward increasing this percentage. To further support transparency, Climate TRACE provides quantitative uncertainty metrics for each estimate upon request, enabling data consumers to make informed judgements about accuracy.

**Table 3**. Total emissions associated with each Climate TRACE sector, share of global total emissions, and share of emissions or activity estimated with peer-reviewed methodologies. Emissions totals and shares are based on 2024, Climate TRACE dataset version 5.4.1.

| **Climate TRACE sector** | **Total sector emissions, $CO_2$e 100yr (Mt)** | **Share of total global emissions (%)** | **Share of emissions estimated with peer-reviewed methodology (%)** |
|---|---|---|---|
| Power | 15761.6 | 20.7 | 99.4 |
| Forestry and Land Use | 15745.6 | 20.7 | 99.8 |
| Manufacturing | 10378.3 | 13.6 | 36.9 |
| Fossil Fuel Operations | 9711.4 | 12.8 | 78 |
| Transportation | 9183.4 | 12.1 | 76.4 |
| Agriculture | 7318.9 | 9.6 | 69.8 |
| Buildings | 4029.1 | 5.3 | 100* |
| Waste | 2078.7 | 2.7 | 40.9 |
| Fluorinated Gases | 1652.5 | 2.2 | 100 |
| Mineral Extraction | 215.5 | 0.3 | 18.9 |
| Total | 76075.0 | 100 | 79.9 |

*At the time of writing, this methodology has been accepted for publication.

### 3.2 Comprehensiveness

By expanding emissions data coverage across space, time, sectors, and pollutants (Figure 2), Climate TRACE consolidates previously fragmented emissions information into a single, accessible dataset [11, 40, 41]. The dataset spans all anthropogenic sectors defined by the IPCC [41] (see Supplementary Dataset 1) and includes all major GHGs, along with key non-GHG pollutants, offering more comprehensive sectoral and pollutant coverage than most individual datasets (Table 2). The inclusion of non-GHG pollutants expands the dataset's scope beyond climate impacts, while asset-level detail enables an improved understanding of pollutant exposure patterns that contribute to global mortality. Additionally, Climate TRACE offers a more detailed sectoral breakdown than most other datasets, enabling more effective mitigation planning. For example, it separately estimates emissions from different manufactured goods that are combined in datasets like EDGAR v8.0 [42].

Climate TRACE provides full global coverage and includes comprehensive data for countries that are underrepresented in datasets like the UNFCCC, which is widely used to track compliance with regulatory frameworks. For example, Climate TRACE provides detailed emissions estimates for major emitters like Iran, Algeria, and Qatar, whose last UNFCCC submissions date back to 2000, 2000, and 2007, respectively [25]. Furthermore, Climate TRACE data include emissions estimates for all 44 countries identified by the United Nations as having the lowest adaptive capacity to climate change [25], many of which lack the resources to develop emissions inventories, empowering least-developed countries to identify and implement feasible mitigation strategies without major new resource investments.

The dataset provides country-level emissions from 2015 to the present and asset-level emissions from 2021 to the present, both with a two-month reporting lag and monthly updates beginning in 2025. This level of temporal detail is especially valuable given that, despite some datasets extending back to the mid-20th century, few (only CAMS, Carbon Monitor, and GRACED) offer data as recent as 2025 (Table 2).

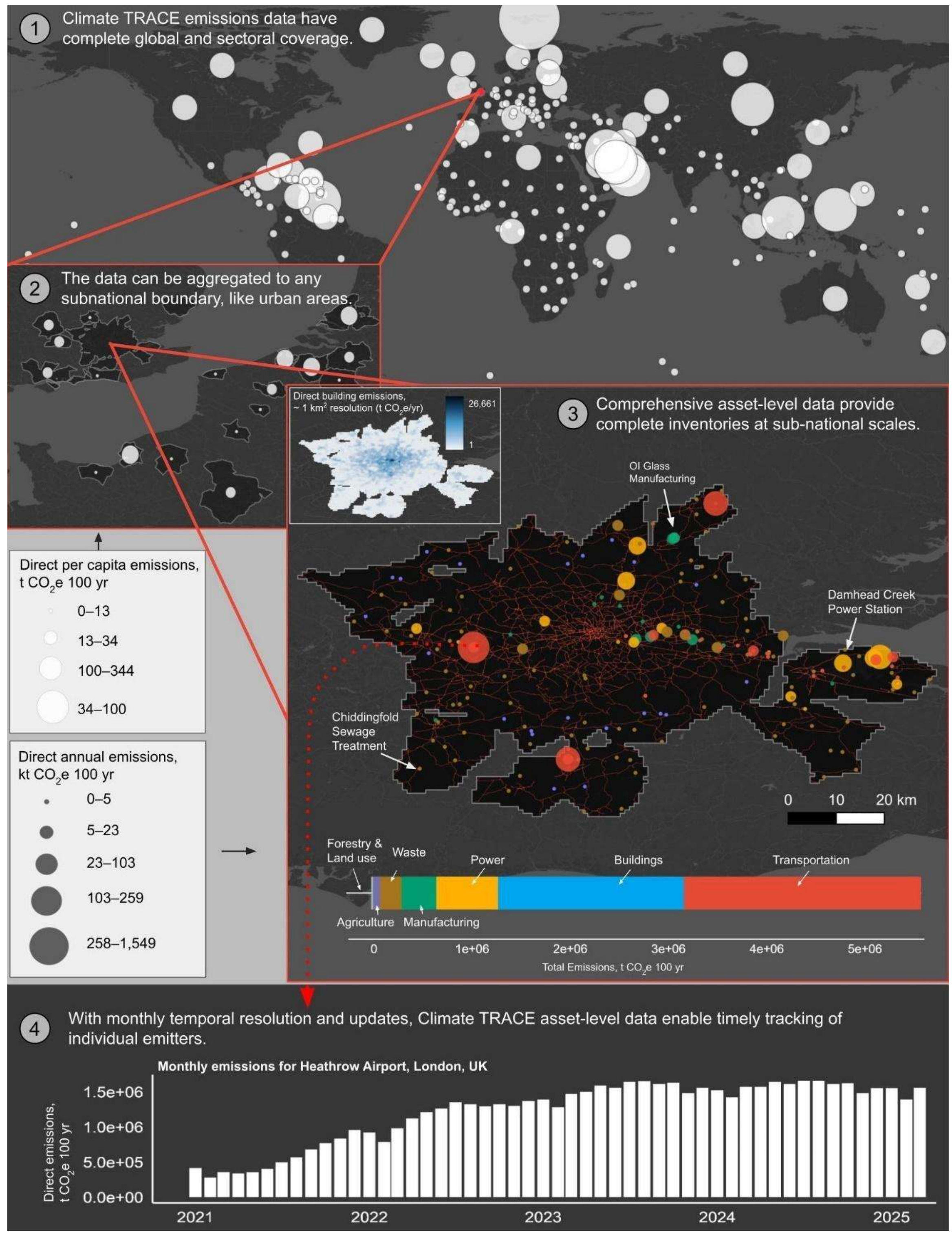

**Figure 1. Map of Climate TRACE features.** Map illustrating the Climate TRACE dataset's comprehensive features, including (1) full global coverage, along with emissions estimates resolved at (2) the London urban area level and (3) individual

emissions sources in London for years 2021 to 2025 (4), with monthly temporal resolution and updates, as shown for Heathrow Airport.

### 3.3 Spatial and temporal resolution

Climate TRACE's monthly facility-level estimates cover 74% of total global anthropogenic emissions, offering the broadest coverage among datasets with this level of detail [31, 32, 33, 41] (Table 2 and Figure 1). While most global datasets provide emissions at the country-level or in gridded formats (Table 2), Climate TRACE data are available at several additional spatial scales including state/province, county, and large urban areas (Table 4). These geographies collectively cover 100% of countries, states, counties, and large urban areas worldwide. At each of these resolutions, 100% of global emissions are estimated, based on IPCC and UNFCCC reporting (25, 43). This eliminates the need for users to generate subnational emissions inventories themselves and offers comprehensive estimates for nearly 60,000 governing units (Table 4). Among these are 260 functional urban areas across the United States, covering all of the country's most populous cities. Despite their critical role in emissions mitigation, few major United States cities had established emissions datasets or reduction targets as recently as 2020 [44]. By filling this gap, cities can access data required for establishing baselines, setting reduction targets, and tracking progress over time.

**Table 4**. **Administrative boundaries**. Administrative boundaries to which Climate TRACE data are aggregated.

| Administrative boundary | Features included in Climate TRACE dataset |
|---|---:|
| Country (**GADM [45] level 0) | 251 |
| State/province (GADM level 1) | 3,660 |
| County (GADM level 2) | 47,218 |
| Urban area (††GHS-FUA [46]) | 9,031 |

**GADM: Global Administrative Areas database.
††GHS-FUA: Global Human Settlement-Functional Urban Areas

The combination of asset-level granularity and global coverage allows for the identification of high-emitting facilities worldwide, including those that may be under-reporting emissions. Climate TRACE reported emissions more than three times higher than those disclosed by several self-reporting companies, with discrepancies especially widespread in the energy sector [47]. The dataset's global coverage and enhanced detail also support global supply chain decarbonization efforts by improving Scope 3 (i.e., value chain) emissions accounting. For example, Climate TRACE data are integrated into Altana (https://altana.ai/resources/carbon-emissions-data), a value chain management platform, replacing country-level emissions with facility-specific estimates. This allows companies to implement interventions that are more tailored to their operations.

For several emissions subsectors in the Climate TRACE dataset, gridded data are presented at resolutions higher than those offered by most other global datasets, including 1 $km^2$ for buildings and land use at 10 $m^2$ resolution for rice cultivation. For building emissions, this represents an 11-fold increase in spatial granularity at the equator compared to most existing gridded datasets. Climate TRACE's monthly temporal resolution and frequent updates, with only a two-month data latency, are more current than most bottom-up datasets, which generally have annual temporal resolution and are updated every one to three years (Table 2). Enhanced resolution and frequent updates support more timely progress tracking and strategy refinement, which stakeholders have identified as critical for decision-making [11, 36].

### 3.4 Usability

Whereas most individual datasets lack a user-friendly visualization platform, Climate TRACE provides an intuitive user interface designed to enhance accessibility, regardless of technical expertise. Through an interactive map and data explorer, stakeholders can access free emissions data to guide mitigation strategies and track progress. Users can search for or navigate to individual facilities or administrative boundaries (countries, states/provinces, counties, urban areas), view emissions aggregated by sector at each level (as shown in Figure 2), and rank assets by emissions totals at a resolution of choice. The interactive map also provides emissions time series for those geographic units. Annual emissions data (2015–2026) can be downloaded in tabular format, aggregated by country, subsector, and type of gas. Monthly asset-level emissions data (2021–2026) can be downloaded by subsector and type of gas. Users can also manually aggregate asset-level data by administrative boundary, with asset locations available for download in geospatial format upon request. Detailed methodology documentation for each emissions subsector is openly available at github.com/climatetracecoalition/methodology-documents, enhancing transparency and ensuring reproducibility, features that are widely recognized as essential for climate action [39].

## 4. Discussion

Climate TRACE addresses a critical gap in the global emissions data landscape by leveraging the strengths of existing datasets and combining them into a single, openly accessible source. The resulting dataset offers improved accuracy, coverage, resolution, and timeliness. Through these enhanced dataset features, the Climate TRACE framework strengthens the infrastructure for emissions verification and climate modeling in ways that existing datasets cannot. Asset-level estimates support the refinement of atmospheric inversion models, a critical approach for measuring surface GHG fluxes. For example, Climate TRACE cattle livestock emissions have been found to align more closely with posterior $CH_4$ flux estimates than coarser-resolution priors [48], underscoring their potential to reconcile discrepancies between estimation methods as coverage expands. Beyond verification, the dataset has potential to enhance our understanding of critical ecosystem processes that are vulnerable to climate change. For example, da Costa and others leveraged the dataset's enhanced coverage and detail to identify when and where forests in Brazil act as a net $CO_2$ sink [49].

In addition to advancing emissions accounting and mitigation efforts, Climate TRACE's consistent global coverage and timely updates reveal emerging global patterns, some of which highlight imbalances in emissions mitigation progress that vary with economic development. For example, between 2015 and 2024, $CO_2$e (100-yr) emissions from Annex I (industrialized economies) countries decreased by an average of 0.26% per year, while emissions from non-Annex I (emerging economies) countries increased by an average of 1.78% per year (Supplementary Figure 1). These findings are consistent with patterns observed in recent decades, where $CO_2$ emissions from low- and middle-income countries increased faster than those from developed countries [50, 44]. Our data also indicate that emissions inequality is growing among cities. Our analysis of 500 major urban areas shows a divergence in mitigation progress between 2021 and 2024: 60.7% of lower-GDP cities saw emissions rise, compared to just 46.5% of higher-GDP cities. On average, emissions increases in lower-GDP cities during this period were 1.5 times higher than those of higher-GDP cities. Section 5 of the Supplementary Materials includes additional global trends.

Despite these advances, we note limitations of the current Climate TRACE dataset that reflect challenges shared across many global emissions datasets. Although the framework seeks to maximize accuracy through the accuracy-based data selection and integration process, the accuracy of model components and emissions estimates cannot be assessed directly in all cases due to the limited availability of in situ emissions measurements. Consequently, some data are incorporated based on proxies for accuracy, leading to variable accuracy of estimates across sectors, regions, and time. This variability across sectors is also impacted by the use of subsector-specific estimation methodologies, as well as the quality and specificity of model inputs. For example, models trained on diverse, independently verified data generally yield more accurate estimates, whereas those that rely on self-reported data or that cannot be verified independently may introduce greater uncertainty. Estimates should, therefore, be interpreted in the context of their uncertainty (which Climate TRACE provides for all emissions data) to ensure appropriate use for climate action planning.

We have introduced Climate TRACE, a framework for building on and extending the strengths of existing data to produce a harmonized global GHG emissions dataset with unprecedented detail and coverage. The framework integrates existing emissions data based on estimated accuracy, filling critical data gaps with Climate TRACE-derived estimates when necessary. The result is a unified, openly accessible dataset delivered through an accessible platform that enables visualization and downloads without the need for data analysis expertise. Climate TRACE is the first global dataset to provide asset-level emissions estimates for all emitting sectors, along with national, provincial, county, and city-level spatial resolution, scales at which decisions are made and policies are implemented. Frequent, timely updates to the dataset enable near-real-time progress tracking toward emissions targets, empowering stakeholders worldwide to take effective climate mitigation action

## 5. Materials and Methods

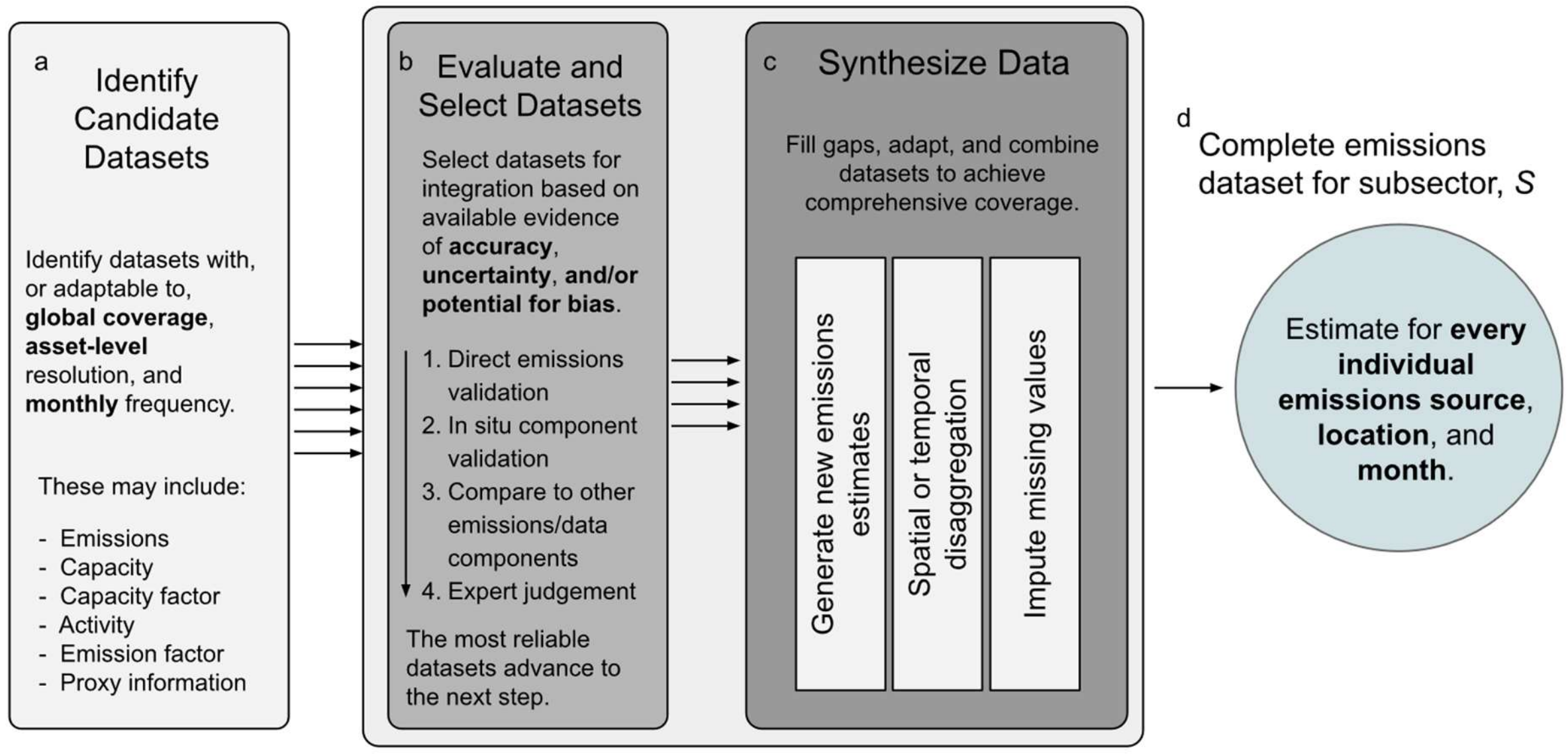

**Figure 2. Climate TRACE's process for achieving complete coverage.** Climate TRACE's process for combining existing datasets and generating new emissions estimates, with the goal of producing a dataset with full global coverage, asset-level spatial resolution and monthly temporal resolution.

### 5.1 Data selection process

For each emissions subsector, Climate TRACE first identifies all datasets that could support emissions estimation and that have, or could be adapted to achieve, (1) full global coverage, (2) asset-level spatial resolution, and (3) monthly temporal resolution (Figure 2a). These candidate datasets may be direct emissions estimates or provide information that can be incorporated into the Climate TRACE framework (see Equations 1 and 2). This includes activity or capacity data that, in combination with emission factors, can yield emissions estimates. Candidate inputs also include proxy information (e.g., production or economic indicators that indirectly reflect emissions).

Each candidate dataset is evaluated for its estimated accuracy, uncertainty, and potential for bias (Figure 2b). Only the datasets with the highest estimated accuracy advance to the next phase. Dataset accuracy can sometimes be quantified directly, such as when dataset providers report uncertainty estimates or when we can independently approximate it by comparing estimates to in situ data. This process is reflected in that approximately 81% of Climate TRACE total global emissions are based on peer-reviewed emissions or activity estimation methodologies (Table 3 and Supplementary Dataset 2). When direct measurements are unavailable, proxies such as activity data or other model inputs can serve as the basis for component-level accuracy evaluation (see section 5.3). In other cases, accuracy is assessed using expert judgment (i.e., the application of specialized knowledge and experience by subject matter experts).

### 5.2 Data synthesis

For each emissions subsector, the selected dataset(s) are then adapted or combined to achieve complete global coverage, asset-level spatial resolution, and monthly temporal resolution (Figure 2c). In addition to synthesizing existing datasets, large information gaps that arise from insufficient resolution or incomplete spatial coverage are addressed by generating new emissions estimates derived from available inputs, effectively building on their individual strengths (see section 5.2.2).

#### 5.2.1 Adapting or combining existing datasets

The coverage and resolution of input datasets is improved by (1) distributing coarser-resolution data across individual facilities, (2) temporally disaggregating coarser-resolution data to monthly resolution, and (3) imputing missing values to ensure a complete and continuous timeseries (see Supplementary Note 1 for detailed descriptions of these protocols). The result is a comprehensive dataset for each subsector, with an emissions estimate for every individual asset, for every location, and for every month (Figure 2d).

#### 5.2.2 Generating new emissions estimates

In some instances, we cannot reach our coverage and resolution goals using existing inputs, in which case we create new emissions estimates (Figure 2c). The Climate TRACE Coalition, consisting of 10 founding organizations and ~140 additional contributors, provides the foundation for this work. Members contribute subsector-specific expertise to produce comprehensive datasets by developing novel methods [51, 48, 49, 52–64] or adapting established approaches.

The emissions estimation process for all Climate TRACE subsectors approximates Equations 1 and 2, which can be expressed as follows for a given emissions subsector s, individual emissions asset or location i, and greenhouse gas g:

$$A_{s,i} = (C_{s,i})\,(CF_{s,i}) \qquad \text{(Equation 1)}$$

$$E_{s,i,g} = (A_{s,i})\,(EF_{s,i,g}) \qquad \text{(Equation 2)}$$

Where $A_{s,i}$ represents activity, calculated as the product of capacity $C_{s,i}$ and capacity factor $CF_{s,i}$. Emissions $E_{s,i,g}$ are then estimated as the product of $A_{s,i}$ and the emission factor $EF_{s,i,g}$. Activity captures the processes or outputs that generate emissions. Capacity is the maximum potential output of a facility under ideal conditions. The capacity factor adjusts this value to reflect actual output, accounting for real-world conditions like maintenance or fluctuating demand. Emission factors quantify the amount of emissions generated per unit of activity and are specific to pollutant, fuel, and/or technology type.

Approaches for estimating the components of Equations 1 and 2 vary by subsector and include machine learning (ML) models that identify assets or predict activity from satellite imagery, globally scalable models linking remote sensing signals to known activity, and direct use of existing datasets (see Supplementary Dataset 2 and Supplementary Note 2 for details). The resulting emissions estimates include $CO_2$, $CO_2$-equivalent ($CO_2e$), $CH_4$, $N_2O$, and non-GHGs, including particulate matter, nitrogen oxides, and sulfur dioxide. Some non-GHG emissions for a few subsectors are estimated using emission factors that account for asset-level information, such as fuel mixes. For the remaining non-GHG estimates, we use a co-pollution method at the country level (see Supplementary Note 3 for details). Some subsectors estimate C, CF, and EF directly and combine them to calculate E. Others estimate total emissions first and subsequently disaggregate emissions to infer capacity, capacity factor, and emission factor. Our sector-specific methods generally incorporate emission factors sourced from the IPCC and the scientific literature, varying by subsector (see Supplementary Dataset 2).

### 5.2.3 Iterative integration

We repeat our data generation process periodically over time to incorporate the latest available data, with the aim of providing the most accurate and complete GHG emissions dataset that can be synthesized from current sources. The Climate TRACE framework is evolving toward a more comprehensive Bayesian approach that would explicitly integrate candidate inputs within a unified probabilistic framework.

### 5.3 Accuracy evaluation framework

Emissions for most Climate TRACE subsectors are not directly observable across all assets, as current satellite monitoring capabilities are limited, and deploying in situ monitors globally is not feasible. Even where direct observations are possible, measurements are typically confined to small, non-representative geographic areas, limiting their usefulness for assessing the accuracy of global emissions estimates. When they can be observed directly, emissions are typically available at small, non-representative geographic scales. For example, GHG emissions from some smaller sources, like dairy farms and rice fields, are below the detection limit of some GHG monitoring satellites (e.g., OCO-2/-3, GHGSat) [65]. As a result, only a limited number of sub-sectors within the Climate TRACE framework can assess emissions accuracy by directly comparing against in situ observations (see Supplementary Dataset 2 for details on accuracy evaluation approaches by emissions sector). Though we try to minimize such occurrences, in some cases, neither in situ measurements nor viable proxies for accuracy assessment are available. The power sub-sector, which accounted for 17.6% of total global emissions in 2024, is a notable exception. An asset-level external accuracy assessment is conducted using in situ emissions data from the US, Europe, Australia, and India [64]. At the asset-level, annual $CO_2$ estimates for this sub-sector have an average error of 1.25 $MtCO_2$ across 7163 assets, with estimates ranging from 0-35 $MtCO_2$.

In the absence of direct observations, most other sub-sectors within the framework employ proxy approaches to assess the accuracy of existing datasets or newly derived estimates, including a component-level accuracy assessment and a comparative evaluation. The component-level accuracy assessment may target any component of Equations 1 and 2, comparing it against spatially or temporally limited in situ observations. Examples include our ML models for road transportation emissions estimation, which are cross-validated against United States in situ average daily traffic measurements collected by roadside devices [59], and our globally scalable models for estimating power generation (i.e., activity for the power sector), which are assessed for accuracy against facility-level in situ power generation data from the United States, Europe, and Australia [54, 62]. This provides an indirect assessment of emissions accuracy by constraining uncertainty in the model inputs, thereby bounding the potential error in the resulting emissions estimates. The comparative evaluation involves evaluating estimates against those from existing datasets, though these reference datasets themselves may lack independent accuracy assessment. This approach bounds error at the output stage by checking for divergence from other datasets, even if none are considered definitive ground truth. Examples include emissions from synthetic fertilizer application [66], industrial wastewater treatment and discharge [67], and oil and gas transport [68].

Although it is not possible to definitively demonstrate that our integrated estimates are more accurate than any individual dataset, our framework enables a dynamic assessment of accuracy in the integrated estimates and continuous incorporation of new information. This increases the likelihood that the resulting estimates are more accurate than any individual dataset. A general description of our data accuracy assessment approach is described in Supplementary Note 4, and sector-specific accuracy assessment approaches are included in Supplementary Dataset 2.

**Author contributions:**
Conceptualization: All authors
Methodology: All authors
Data Curation: All authors
Writing—Review & Editing: All authors
Writing—Original Draft Preparation: BVL, JMM, KB, AD, GM, TMG PJM
Visualization: BVL, JMM, KB, AD, GM, TMG PJM

**Competing interests:** The authors declare they have no competing interests.

**Data and materials availability:** All data underlying this study are available in the Supplementary Materials, on the Climate TRACE website, or on a data repository. All Climate TRACE emissions data, including the source data for Figure 1, are available for download from the Climate TRACE website, https://climatetrace.org/explore. The data supporting the analyses presented in the discussion section, along with the R scripts used to reproduce the results, are publicly available in Zenodo with the identifier https://doi.org/10.5281/zenodo.16944396.

**References**

1. V. Masson-Delmotte, H.-O. Pörtner, J. Skea, P. Zhai, D. Roberts, P. R. Shukla, A. Pirani, R. Pidcock, Y. Chen, E. Lonnoy, W. Moufouma-Okia, C. Péan, S. Connors, J. B. R. Matthews, X. Zhou, M. I. Gomis, T. Maycock, M. Tignor, T. Waterfield, An IPCC Special Report on the impacts of global warming of 1.5°C above pre-industrial levels and related global greenhouse gas emission pathways, in the context of strengthening the global response to the threat of climate change, sustainable development, and efforts to eradicate poverty.

2. M. Roelfsema, H. Fekete, N. Höhne, M. Den Elzen, N. Forsell, T. Kuramochi, H. De Coninck, D. P. Van Vuuren, Reducing global GHG emissions by replicating successful sector examples: the 'good practice policies' scenario. Clim. Policy **18**, 1103–1113 (2018).

3. D. Bastviken, J. Wilk, N. T. Duc, M. Gålfalk, M. Karlson, T.-S. Neset, T. Opach, A. Enrich-Prast, I. Sundgren, Critical method needs in measuring greenhouse gas fluxes. Environ. Res. Lett. **17**, 104009 (2022).

4. L. Perugini, G. Pellis, G. Grassi, P. Ciais, H. Dolman, J. I. House, G. P. Peters, P. Smith, D. Günther, P. Peylin, Emerging reporting and verification needs under the Paris Agreement: How can the research community effectively contribute? Environ. Sci. Policy **122**, 116–126 (2021).

5. K. Rypdal, W. Winiwarter, Uncertainties in greenhouse gas emission inventories — evaluation, comparability and implications. Environ. Sci. Policy **4**, 107–116 (2001).

6. L. Yona, Emissions Omissions: Greenhouse Gas Accounting Gaps. Harv. Environ. LAW Rev. (2025).

7. M. R. Boswell, A. I. Greve, T. L. Seale, An Assessment of the Link Between Greenhouse Gas Emissions Inventories and Climate Action Plans. J. Am. Plann. Assoc. **76**, 451–462 (2010).

8. T. Oda, S. Maksyutov, A very high-resolution (1 km×1 km) global fossil fuel $CO_2$ emission inventory derived using a point source database and satellite observations of nighttime lights. Atmospheric Chem. Phys. **11**, 543–556 (2011).

9. C. Yaman, A Review on the Process of Greenhouse Gas Inventory Preparation and Proposed Mitigation Measures for Reducing Carbon Footprint. Gases **4**, 18–40 (2024).

10. United Nations Framework Convention on Climate Change (UNFCCC). Paris Agreement. UNFCCC (2015). https://unfccc.int/sites/default/files/resource/parisagreement_publication.pdf

11. National Academies of Sciences, Engineering, and Medicine. Greenhouse Gas Emissions Information for Decision Making: A Framework Going Forward. Washington, DC: The National Academies Press, 2022. https://doi.org/10.17226/26641.

12. R. Swart, P. Bergamaschi, T. Pulles, F. Raes, Are national greenhouse gas emissions reports scientifically valid? Clim. Policy **7**, 535–538 (2007).

13. K. M. Dittmer, E. Wollenberg, M. Cohen, C. Egler, How good is the data for tracking countries' agricultural greenhouse gas emissions? Making use of multiple national greenhouse gas inventories. Front. Sustain. Food Syst. **7** (2023).

14. K. Gurney, P. Shepson, The power and promise of improved climate data infrastructure. Proc. Natl. Acad. Sci. **118**, e2114115118 (2021).

15. Macknick, J., & Grubler, A. Energy and Carbon Dioxide Emission Data Uncertainties. International Institute for Applied Systems Analysis (IIASA) Working Paper IR-09-032 (2009). https://pure.iiasa.ac.at/id/eprint/9119/1/IR-09-032.pdf. Accessed 22 April 2026.

16. N. Milojevic-Dupont, F. Creutzig, Machine learning for geographically differentiated climate change mitigation in urban areas. Sustain. Cities Soc. **64**, 102526 (2021).

17. L. Yona, B. Cashore, R. B. Jackson, J. Ometto, M. A. Bradford, Refining national greenhouse gas inventories. Ambio **49**, 1581–1586 (2020).

18. R. M. Andrew, A comparison of estimates of global carbon dioxide emissions from fossil carbon sources. Earth Syst. Sci. Data **12**, 1437–1465 (2020).

19. J. Macknick, Energy and CO2 emission data uncertainties. Carbon Manag. **2**, 189–205 (2011).

20. G. Marland, A. Brenkert, J. Olivier, CO2 from fossil fuel burning: a comparison of ORNL and EDGAR estimates of national emissions. Environ. Sci. Policy **2**, 265–273 (1999).

21. A. van Amstel, J. Olivier, L. Janssen, Analysis of differences between national inventories and an Emissions Database for Global Atmospheric Research (EDGAR). Environ. Sci. Policy **2**, 275–293 (1999).

22. M. Crippa, D. Guizzardi, F. Pagani, M. Schiavina, M. Melchiorri, E. Pisoni, F. Graziosi, M. Muntean, J. Maes, L. Dijkstra, M. Van Damme, L. Clarisse, P. Coheur, Insights on the spatial distribution of global, national and sub-national GHG emissions in EDGARv8.0. (2024). https://doi.org/10.5194/essd-2023-514.

23. E. E. McDuffie, S. J. Smith, P. O'Rourke, K. Tibrewal, C. Venkataraman, E. A. Marais, B. Zheng, M. Crippa, M. Brauer, R. V. Martin, A global anthropogenic emission inventory of atmospheric pollutants from sector- and fuel-specific sources (1970–2017): an application of the Community Emissions Data System (CEDS). Earth Syst. Sci. Data **12**, 3413–3442 (2020).

24. International Energy Agency (IEA), "IEA CO2 Emissions in 2022" (2023).

25. UNFCCC, UNFCCC Greenhouse Gas Inventory Data - Detailed data by Party; https://di.unfccc.int/detailed_data_by_party?_gl=1*kr0f9o*_ga*MTY2NjQ5OTQyMy4xNzIxNjU3Mjk5*_ga_7ZZWT14N79*MTczMzI0NzE5Ny4zOC4wLjE3MzMyNDcyMDAuMC4wLjA. Accessed22 April 2026.

26. Liu, Z., Ciais, P., Deng, Z., Davis, S. J., Zheng, B., Wang, Y., et al. (2020). Carbon Monitor, a near-real-time daily dataset of global CO2 emission from fossil fuel and cement production. Scientific data, 7(1), 392.

27. Gilfillan, D., & Marland, G. (2021). CDIAC-FF: global and national CO2 emissions from fossil fuel combustion and cement manufacture: 1751–2017. Earth System Science Data, 13(4), 1667-1680.

28. Friedlingstein, P., O'Sullivan, M., Jones, M. W., Andrew, R. M., Hauck, J., Landschützer, P., et al. (2024). Global carbon budget 2024. Earth System Science Data Discussions, 2024, 1-133.

29. National Oceanic and Atmospheric Administration (NOAA). CarbonTracker Documentation CT2022 release. https://gml.noaa.gov/ccgg/carbontracker/documentation.php. Accessed 22 April 2026 (2023).

30. Nations Are Undercounting Emissions, Putting UN Goals at Risk. Yale E360 (2024). https://e360.yale.edu/features/undercounted-emissions-un-climate-change. Accessed22 April 2026.

31. Global Infrastructure Emissions Detector (GID). Emissions Inventory (2021). http://gidmodel.org.cn/?page_id=1425. Accessed 22 April 2026.

32. R. Bun, Z. Nahorski, J. Horabik-Pyzel, O. Danylo, L. See, N. Charkovska, P. Topylko, M. Halushchak, M. Lesiv, M. Valakh, V. Kinakh, Development of a high-resolution spatial inventory of greenhouse gas emissions for Poland from stationary and mobile sources. Mitig. Adapt. Strateg. Glob. Change **24**, 853–880 (2019).

33. A. Kato, K. R. Gurney, G. S. Roest, P. Dass, Exploring differences in $FFCO_2$ emissions in the United States: comparison of the Vulcan data product and the EPA national GHG inventory. Environ. Res. Lett. **18**, 124043 (2023).

34. Gurney, K. Fossil Fuel Data Assimilation System (FFDAS) Data. https://ffdas.rc.nau.edu/Data.html. Accessed 22 April 2026.

35. M. W. Jones, R. M. Andrew, G. P. Peters, G. Janssens-Maenhout, A. J. De-Gol, X. Dou, Z. Liu, P. Pickers, P. Ciais, P. K. Patra, F. Chevallier, C. Le Quéré, Gridded fossil CO2 emissions and related O2 combustion consistent with national inventories, version GCP-GridFEDv2024.0, Zenodo (2024); https://doi.org/10.5281/zenodo.13909046.

36. T. Oda, S. Maksyutov, R. J. Andres, The Open-source Data Inventory for Anthropogenic $CO_2$, version 2016 (ODIAC2016): a global monthly fossil fuel $CO_2$ gridded emissions data product for tracer transport simulations and surface flux inversions. Earth Syst. Sci. Data **10**, 87–107 (2018).

37. Climate Watch. Climate Watch Data Explorer. https://www.climatewatchdata.org/data-explorer (2025). Accessed 22 April 2026.

38. Food and Agriculture Organization Statistics (FAOSTAT). FAOSTAT Emissions Totals (2025). https://www.fao.org/faostat/en/#data/Gt. Accessed 22 April 2026.

39. E. Romijn, V. De Sy, M. Herold, H. Böttcher, R. M. Roman-Cuesta, S. Fritz, D. Schepaschenko, V. Avitabile, D. Gaveau, L. Verchot, C. Martius, Independent data for transparent monitoring of greenhouse gas emissions from the land use sector – What do stakeholders think and need? Environ. Sci. Policy **85**, 101–112 (2018).

40. J. C. Minx, W. F. Lamb, R. M. Andrew, J. G. Canadell, M. Crippa, N. Döbbeling, P. M. Forster, D. Guizzardi, J. Olivier, G. P. Peters, J. Pongratz, A. Reisinger, M. Rigby,

M. Saunois, S. J. Smith, E. Solazzo, H. Tian, A comprehensive and synthetic dataset for global, regional, and national greenhouse gas emissions by sector 1970–2018 with an extension to 2019. Earth Syst. Sci. Data **13**, 5213–5252 (2021).

41. H. S. Eggleston, L. Buendia, K. Miwa, T. Ngara, K. Tanabe, "2006 IPCC guidelines for national greenhouse gas inventories" (2006). https://www.ipcc-nggip.iges.or.jp/public/2006gl/. Accessed 22 April 2026.

42. M. Crippa, D. Guizzardi, F. Pagani, M. Schiavina, M. Melchiorri, E. Pisoni, F. Graziosi, M. Muntean, J. Maes, L. Dijkstra, M. Van Damme, L. Clarisse, P. Coheur, Insights into the spatial distribution of global, national, and subnational greenhouse gas emissions in the Emissions Database for Global Atmospheric Research (EDGAR v8.0). Earth Syst. Sci. Data **16**, 2811–2830 (2024).

43. U.S. Environmental Protection Agency, Greenhouse Gas Reporting Program (GHGRP). https://www.epa.gov/ghgreporting, (2023).

44. S. A. Markolf, Pledges and progress: Steps toward greenhouse gas emissions reductions in the 100 largest cities across the United States. (2020). https://www.brookings.edu/articles/pledges-and-progress-steps-toward-greenhouse-gas-emissions-reductions-in-the-100-largest-cities-across-the-united-states/. Accessed 22 April 2026.

45. GADM: Global Administrative Areas database, (2024); https://gadm.org/.

46. GHSL data package 2019, GHS functional urban areas (GHS-FUA), derived from Sentinel data, R2019A; https://human-settlement.emergency.copernicus.eu/.

47. M. Lepere, S. Maso, Y. Dong, D. Aikman, Emissions dissonance: Examining how firm-level under-reporting undermines policy (2025). https://d1e00ek4ebabms.cloudfront.net/production/uploaded-files/Lepere_Emissions_Dissonance_SELE2025-35ef4e3f-465b-46a7-81e9-02e562d99575.pdf. Accessed22 April 2026.

48. S. E. Hancock, D. J. Jacob, Z. Chen, H. Nesser, A. Davitt, D. J. Varon, M. P. Sulprizio, N. Balasus, L. A. Estrada, M. Cazorla, L. Dawidowski, S. Diez, J. D. East, E. Penn, C. A. Randles, J. Worden, I. Aben, R. J. Parker, J. D. Maasakkers, Satellite quantification of methane emissions from South American countries: a high-resolution inversion of TROPOMI and GOSAT observations. Atmospheric Chem. Phys. **25**, 797–817 (2025).

49. L. M. da Costa, A. Davitt, G. Volpato, G. C. de Mendonça, A. R. Panosso, N. La Scala, A comparative analysis of GHG inventories and ecosystems carbon absorption in Brazil. Sci. Total Environ. **958**, 177932 (2025).

50. The Intergovernmental Panel on Climate Change (IPCC) Data. https://www.ipcc.ch/data/. Accessed 22 April 2026.

51. D. Saha, B. Basso, G. P. Robertson, Machine learning improves predictions of agricultural nitrous oxide (N2O) emissions from intensively managed cropping systems. Environ. Res. Lett. **16**, 024004 (2021).

52. F. I. Ginting, R. Rudiyanto, Fatchurrachman, R. Mohd Shah, N. Che Soh, S. G. Eng Giap, D. Fiantis, B. I. Setiawan, S. Schiller, A. Davitt, B. Minasny, High-resolution maps of rice cropping intensity across Southeast Asia. Sci. Data **12**, 1408 (2025).

53. Fatchurrachman, Rudiyanto, N. C. Soh, R. M. Shah, S. G. E. Giap, B. I. Setiawan, B. Minasny, High-Resolution Mapping of Paddy Rice Extent and Growth Stages across Peninsular Malaysia Using a Fusion of Sentinel-1 and 2 Time Series Data in Google Earth Engine. Remote Sens. **14**, 1875 (2022).

54. M. Hobbs, A. R. Kargar, H. Couture, J. Freeman, I. Söldner-Rembold, A. Ferreira, J. Jeyaratnam, J. O'Connor, J. Lewis, H. Koenig, C. McCormick, T. Nakano, C. Dalisay, A. Davitt, L. Gans, C. Lewis, G. Volpato, M. Gray, G. McCormick, "Inferring Carbon Dioxide Emissions From Power Plants Using Satellite Imagery and Machine Learning" in IGARSS 2023 - 2023 IEEE International Geoscience and Remote Sensing Symposium (2023). https://ieeexplore.ieee.org/document/10283046), 4911–4914.

55. C. Kruse, E. Boyda, S. Chen, K. Karra, T. Bou-Nahra, D. Hammer, J. Mathis, T. Maddalene, J. Jambeck, F. Laurier, Satellite monitoring of terrestrial plastic waste. PLOS ONE **18**, e0278997 (2023).

56. X. Lu, D. J. Jacob, H. Wang, J. D. Maasakkers, Y. Zhang, T. R. Scarpelli, L. Shen, Z. Qu, M. P. Sulprizio, H. Nesser, A. A. Bloom, S. Ma, J. R. Worden, S. Fan, R. J. Parker, H. Boesch, R. Gautam, D. Gordon, M. D. Moran, F. Reuland, C. A. O. Villasana, A. Andrews, Methane emissions in the United States, Canada, and Mexico: evaluation of national methane emission inventories and 2010–2017 sectoral trends by inverse analysis of in situ (GLOBALVIEWplus $CH_4$ ObsPack) and satellite (GOSAT) atmospheric observations. Atmospheric Chem. Phys. **22**, 395–418 (2022).

57. R. Mukherjee, D. Rollend, G. Christie, A. Hadzic, S. Matson, A. Saksena, M. Hughes, "Towards Indirect Top-Down Road Transport Emissions Estimation" (2021). https://openaccess.thecvf.com/content/CVPR2021W/EarthVision/papers/Mukherjee_Towards_Indirect_Top-Down_Road_Transport_Emissions_Estimation_CVPRW_2021_paper.pdf, 1092–1101. Accessed 22 April 2026.

58. D. L. Northrup, B. Basso, M. Q. Wang, C. L. S. Morgan, P. N. Benfey, Novel technologies for emission reduction complement conservation agriculture to achieve negative emissions from row-crop production. Proc. Natl. Acad. Sci. **118**, e2022666118 (2021).

59. D. Rollend, K. Foster, T. M. Kott, R. Mocharla, R. Muñoz, N. Fendley, C. Ashcraft, F. Willard, E. P. Reilly, M. Hughes, Machine learning for activity-based road transportation emissions estimation. Environ. Data Sci. **2**, e38 (2023).

60. Rudiyanto, B. Minasny, R. Shah, N. Che Soh, C. Arif, B. Indra Setiawan, Automated Near-Real-Time Mapping and Monitoring of Rice Extent, Cropping Patterns, and Growth Stages in Southeast Asia Using Sentinel-1 Time Series on a Google Earth Engine Platform. Remote Sens. **11**, 1666 (2019).

61. T. R. Scarpelli, D. J. Jacob, S. Grossman, X. Lu, Z. Qu, M. P. Sulprizio, Y. Zhang, F. Reuland, D. Gordon, J. R. Worden, Updated Global Fuel Exploitation Inventory (GFEI) for methane emissions from the oil, gas, and coal sectors: evaluation with inversions of atmospheric methane observations. Atmospheric Chem. Phys. **22**, 3235–3249 (2022).

62. Strong, B., Boyda, E., Kruse, C., Ingold, T. & Maron, M. Digital Applications Unlock Remote Sensing AI Foundation Models for Scalable Environmental Monitoring. Frontiers in Climate. **7** (2025).

63. L. Xu, S. S. Saatchi, Y. Yang, Y. Yu, J. Pongratz, A. A. Bloom, K. Bowman, J. Worden, J. Liu, Y. Yin, G. Domke, R. E. McRoberts, C. Woodall, G.-J. Nabuurs, S. de-Miguel, M. Keller, N. Harris, S. Maxwell, D. Schimel, Changes in global terrestrial live biomass over the 21st century. Sci. Adv. **7**, eabe9829 (2021).

64. H. D. Couture, M. Alvara, J. Freeman, A. Davitt, H. Koenig, A. Rouzbeh Kargar, J. O'Connor, I. Söldner-Rembold, A. Ferreira, J. Jeyaratnam, J. Lewis, C. McCormick, T. Nakano, C. Dalisay, C. Lewis, G. Volpato, M. Gray, G. McCormick, Estimating Carbon Dioxide Emissions from Power Plant Water Vapor Plumes Using Satellite Imagery and Machine Learning. Remote Sens. **16**, 1290 (2024).

65. O. Moeini, R. Nassar, J.-P. Mastrogiacomo, M. Dawson, C. W. O'Dell, R. R. Nelson, A. Chatterjee, Quantifying CO2 Emissions From Smaller Anthropogenic Point Sources Using OCO-2 Target and OCO-3 Snapshot Area Mapping Mode Observations. J. Geophys. Res. Atmospheres **130**, e2024JD042333 (2025).

66. P. Sharma, B. Basso, Agriculture sector: Emission from Synthetic Fertilizer Application. (2025). https://github.com/climatetracecoalition/methodology-documents/blob/main/2025/Agriculture/Synthetic%20Fertilizer%20Crop%20Residue%20and%20Manure%20Application%20Emissions-AgricultureSector-021926.pdf. Accessed 22 April 2026.

67. G. Collins, A. Jain, L. Sridhar, E. Reilly, Waste sector: Emissions from Wastewater Treatment Plants. (2025). https://github.com/climatetracecoalition/methodology-documents/blob/main/2025/Waste/Wastewater%20Treatment%20Plants%20Emissions-WasteSector-112025.docx.pdf. Accessed 22 April 2026.

68. L. Schmeisser, A. Tecza, R. Wang, M. Huffman, S. Schadel, S. Bylsma, J. Hansen, Z. Schmidt, T. Conway, D. Gordon, Fossil Fuel Operations sector: Oil and Gas Production and Transport Emissions. (2025). https://github.com/climatetracecoalition/methodology-documents/blob/main/2025/Fossil%20Fuel%20Operations/Oil%20and%20Gas%20Pr

oduction%20and%20Transport-FossilFuelOperationsSector-112025.docx.pdf. Accessed 22 April 2026.

69. United Nations Industrial Development Organization Statistics Portal (UNIDO). UNIDO INDSTAT Database. https://stat.unido.org/

70. M. Crippa, E. Solazzo, G. Huang, D. Guizzardi, E. Koffi, M. Muntean, C. Schieberle, R. Friedrich, G. Janssens-Maenhout, High resolution temporal profiles in the Emissions Database for Global Atmospheric Research. Sci. Data **7**, 121 (2020).

71. K. Raniga, D. Moore, Z. Doctor, C. Lewis, L. Sridhar, P. Thomas, I. Saraswat, G. Collins, A. Nellis, N. Brown, M. Pekala, E. Reilly, M. Hughes, G. McCormick, Temporal Disaggregation of Emissions Data for the Climate TRACE Inventory (2024). https://github.com/climatetracecoalition/methodology-documents/blob/main/2025/Post%20Processing%20for%20Global%20Emissions%20and%20Metadata%20Completeness/Temporal%20Disaggregation%20of%20Emissions%20Data%20for%20the%20Climate%20TRACE%20Inventory-121225.docx.pdf. Accessed 22 April 2026.

72. D. Rolnick, P. L. Donti, L. H. Kaack, K. Kochanski, A. Lacoste, K. Sankaran, A. S. Ross, N. Milojevic-Dupont, N. Jaques, A. Waldman-Brown, Tackling climate change with machine learning. ArXiv Prepr. ArXiv190605433 (2019).

73. D. H. Cusworth, R. M. Duren, A. K. Thorpe, E. Tseng, D. Thompson, A. Guha, S. Newman, K. T. Foster, C. E. Miller, Using remote sensing to detect, validate, and quantify methane emissions from California solid waste operations. Environ. Res. Lett. **15**, 054012 (2020).

74. H. Bovensmann, M. Buchwitz, J. P. Burrows, M. Reuter, T. Krings, K. Gerilowski, O. Schneising, J. Heymann, A. Tretner, J. Erzinger, A remote sensing technique for global monitoring of power plant CO2; emissions from space and related applications. Atmospheric Meas. Tech. **3**, 781–811 (2010).

75. R. DeFries, F. Achard, S. Brown, M. Herold, D. Murdiyarso, B. Schlamadinger, C. De Souza, Earth observations for estimating greenhouse gas emissions from deforestation in developing countries. Environ. Sci. Policy **10**, 385–394 (2007).

76. C. A. Rotz, Modeling greenhouse gas emissions from dairy farms. J. Dairy Sci. **101**, 6675–6690 (2018).

77. L. A. Harper, T. K. Flesch, J. M. Powell, W. K. Coblentz, W. E. Jokela, N. P. Martin, Ammonia emissions from dairy production in Wisconsin. J. Dairy Sci. **92**, 2326–2337 (2009).

78. N. T. Vechi, J. Mellqvist, C. Scheutz, Quantification of methane emissions from cattle farms, using the tracer gas dispersion method. Agric. Ecosyst. Environ. **330**, 107885 (2022).

79. M. A. Hasan, A. A. Mamun, S. M. Rahman, K. Malik, M. I. U. Al Amran, A. N. Khondaker, O. Reshi, S. P. Tiwari, F. S. Alismail, Climate Change Mitigation Pathways for the Aviation Sector. Sustainability **13**, 3656 (2021).

80. International Civil Aviation Organization (ICAO), "ICAO Carbon Calculator Methodology" (2018); https://icec.icao.int/Documents/Methodology%20ICAO%20Carbon%20Emissions%20Calculator_v13_Final.pdf. Accessed 22 April 2026.

81. M. J. Eckelman, J. D. Sherman, A. J. MacNeill, Life cycle environmental emissions and health damages from the Canadian healthcare system: An economic-environmental-epidemiological analysis. PLOS Med. **15**, e1002623 (2018).

82. United States Environmental Protection Agency (EPA). AP-42: Compilation of Air Emissions Factors from Stationary Sources (2016). https://www.epa.gov/air-emissions-factors-and-quantification/ap-42-compilation-air-emissions-factors-stationary-sources. Accessed 22 April 2026.

83. European Monitoring and Evaluation Programme/European Environment Agency (EMEP/EEA). Air pollutant emission inventory guidebook 2023 (2023). https://www.eea.europa.eu/en/analysis/publications/emep-eea-guidebook-2023. Accessed 22 April 2026.

84. K. Miyazaki, K. Bowman, Predictability of fossil fuel CO2 from air quality emissions. Nat. Commun. **14**, 1604 (2023).

85. R. M. Hoesly, S. J. Smith, L. Feng, Z. Klimont, G. Janssens-Maenhout, T. Pitkanen, J. J. Seibert, L. Vu, R. J. Andres, R. M. Bolt, T. C. Bond, L. Dawidowski, N. Kholod, J. Kurokawa, M. Li, L. Liu, Z. Lu, M. C. P. Moura, P. R. O'Rourke, Q. Zhang, Historical (1750–2014) anthropogenic emissions of reactive gases and aerosols from the Community Emissions Data System (CEDS). Geosci. Model Dev. **11**, 369–408 (2018).

86. M. Crippa, D. Guizzardi, F. Pagani, M. Banja, M. Muntean, E. Schaaf, W. Becker, F. Monforti-Ferrario, R. Quadrelli, A. Risquez Martin, P. Taghavi-Moharamli, J. Koykka, G. Grassi, S. Rossi, J. Brandao De Melo, D. Oom, A. Branco, J. San-Miguel, E. Vignati, "GHG Emissions of all World Countries: 2023" (2023); https://www.nature.com/articles/296283c0.

87. Climate TRACE, Climate TRACE Emissions Data: All Sectors (2026). https://climatetrace.org/explore. Accessed 22 April 2026.

## Supplementary Materials

**The document includes:**

Supplementary Notes 1–5
Supplementary Results
Supplementary Figure 1
Supplementary Table 1
Supplementary Table 2

**Other Supplementary Materials for this manuscript include the following, which can be accessed at https://zenodo.org/records/21139462:**

**Supplementary Dataset 1 (separate spreadsheet)**
Provides each emissions subsector included in the Climate TRACE dataset and the equivalent Intergovernmental Panel on Climate Change (IPCC) sector(s).

**Supplementary Dataset 2 (separate spreadsheet)**
Provides a detailed methodological description of each subsector-specific emissions estimation and validation approach.

## Supplementary Text

Supplementary Note 1.0 Adapting or combining existing datasets

Climate TRACE fills emissions data gaps by adapting or combining the highest-quality available data that have the potential to achieve comprehensive coverage. Data quality is assessed based on available evidence of accuracy, uncertainty, and potential for bias. Where no suitable emissions data exist, Climate TRACE generates entirely new emissions estimates using sector-specific methodologies (see Supplementary Note 2.0).

Supplementary Note 1.1 Data-informed disaggregation approach

In cases where the highest quality or most accurate estimates available have country-level resolution, Climate TRACE adapts those datasets to improve their spatial granularity. Country-level emissions estimates from existing datasets (e.g., EDGAR, CEDS, UNFCCC) are allocated to individual assets according to approximations of activity (hereafter referred to as the "data-informed disaggregation approach" and labelled as "Disaggregation" in Column L of Supplementary Dataset 2). Activity is approximated using relevant, publicly available production information, including asset location, size, known production value, or similar assets in other countries.

Examples of subsectors that use the data-informed disaggregation approach include textile, leather, and apparel manufacturing and food, beverages, and tobacco manufacturing. For these sectors, country-level economic output (in USD) from the United Nations Industrial Development Organization (UNIDO) [*69*] was used as a proxy for production (i.e., activity). Output-per-asset values were calculated by dividing total economic output by the number of emitting establishments and were used as asset-level proxies for activity. To avoid overcounting, adjustments were made to exclude non-emitting facilities (e.g., offices, headquarters). A global ratio of emitting to total assets identified from online or public sources (hereafter referred to as "scraped" assets) was applied to the UNIDO establishment counts to estimate the number of emitting facilities. For example, only facility types like "wet processing" or "tannery" were considered to be "emitting" in textiles-leather-apparel subsector. If the number of scraped emitting assets in a given country exceeded the adjusted UNIDO estimate, the scraped count was used instead.

Some scraped assets had corresponding emissions data, which were extrapolated to complete their time series. For each country, gas, and year, if the sum of emissions from these assets exceeded the country-level total, it replaced the national estimate. To assign emissions to other scraped assets that had activity data but no emissions data, a country-specific emissions factor was calculated by dividing total emissions by total activity (e.g., economic output or production). This factor was then applied to estimate emissions at the asset level. However, if the scraped asset emissions already met or exceeded the country total, the remaining assets' emissions and metadata were imputed using default values (i.e., the median emissions factor, median capacity factor, and mean capacity from other assets in the same country, or global averages if local data were unavailable). More detailed information on this approach, including a comparison of Climate TRACE country-level emissions to EDGAR for sectors subject to the data-informed disaggregation approach can be found on the Climate TRACE GitHub methodology repository: https://github.com/climatetracecoalition/methodology-documents/tree/main in the "Data-Informed Disaggregation and Implicit Estimation of Emissions in Other Subsectors" file within the "Other Sectors" folder.

Supplementary Note 1.2 Spatial disaggregation of spatially uncertain emissions

To ensure spatially complete emissions estimates, Climate TRACE combines three sources of information: (1) Climate TRACE-derived facility-level emissions, (2) country-level estimates from the highest quality datasets available for each subsector (e.g., EDGAR, CEDS, UNFCCC, or aggregated facility-level emissions), and (3) proxy datasets. The goal is to generate accurate emissions totals for each administrative boundary, city, and grid cell. This approach is labelled as "Disaggregation" in Column L of Supplementary Dataset 2.

We begin by assuming that the country-level estimate represents a lower bound on actual emissions. If the Climate TRACE facility-level total for a given country and sector exceeds the country-level estimate, the Climate TRACE-derived facility-level data are assumed to be more complete (i.e., account for emissions that are missing from other datasets) and the most accurate available. The emissions estimated for each individual facility by Climate TRACE are aggregated within defined geographic areas, such as countries, states, cities, or grid cells, to produce total emissions estimates within those administrative boundaries for improved downstream usability. Conversely, if the country-level estimate is higher than the Climate TRACE-derived facility-level total, we assume the total emissions are not fully captured by the facility-level estimates (i.e., some emissions are unaccounted for), and there are some emissions with high spatial uncertainty within the country. In this case, we revert to the country-level estimate as a lower bound and distribute these unaccounted for (i.e., spatially uncertain) emissions across space using sector-specific proxy data (e.g., population, nightlight intensity, industrial activity) aggregated to GADM, city, or grid cell boundaries to match our prior expectation of where emissions originate from. This process preserves the total emissions for a given country and the facility-level estimates, while allocating emissions without specific facility locations in alignment with prior expectations.

Proxying is applied only after subtracting known facility-level emissions, ensuring remainders are allocated only to areas with missing data. Where partial facility data are available, remainder emissions may be further allocated using a data-informed disaggregation approach. It should also be noted that, in some cases, country totals for a given subsector are not based on external datasets but are instead calculated as the sum of emissions from all facilities in that country.

The process for disaggregating spatially uncertain emissions indicates potential gaps in facility-level detection: areas with large amounts of spatially uncertain emissions (i.e., low percentage of emissions allocated to facilities [see Column H of Supplementary Dataset 2]) may signal sectors or geographies where Climate TRACE facility coverage is incomplete and needs improvement. However, we do not assert which source is definitively over- or under-estimating emissions when totals are close; in such cases, uncertainty remains. A summary of Climate TRACE emissions totals and corresponding spatially uncertain emissions can be found in Table 2 of the Spatial Disaggregation of Remainder Emissions document, available in the Climate TRACE GitHub methodology repository: https://github.com/climatetracecoalition/methodology-documents/tree/main.

Supplementary Note 1.3 Subtraction-based estimation approach

The implicit estimation approach is utilized when there is insufficient information to reliably estimate facility-level emissions or to disaggregate coarser resolution emissions to individual assets. In such cases, broader emissions totals from datasets external to Climate TRACE, which often aggregate multiple subsectors, are compared with more specific emissions categories that are already accounted for in Climate TRACE. The difference between the two is used to infer emissions for specific subsectors. For example, we subtract iron, steel, and aluminum emissions from EDGAR's Metal Industry sector to derive our "other metals" emissions estimate. Then, in the absence of sufficiently detailed location data for an asset-level disaggregation, the data-informed disaggregation approach is used to spatially disaggregate emissions. Emissions for most subsectors estimated with this method account for a small (<1%) fraction of the global total (e.g., other mining and quarrying; 0.06%, biological treatment of solid waste; 0.04%, other manure management; 0.38%, incineration and open burning of waste; 0.11%). A few subsectors account for a more substantial share of global emissions (e.g., other manufacturing; 2.89%, fluorinated gases; 2.4%, other energy use; 2.71%, other agricultural soil emissions; 2.34%). More detailed information on this approach can be found on the Climate TRACE GitHub methodology repository: https://github.com/climatetracecoalition/methodology-documents/tree/main in the "Data-Informed Disaggregation and Implicit Estimation of Emissions in Other Subsectors" file within the "Other Sectors" folder.

Supplementary Note 1.4 Temporal disaggregation: Imputation of missing values and standardization to monthly estimates

Prior to analysis, emissions estimates and their associated input data are preprocessed to ensure a complete and consistent timeseries. Missing values are first resolved at the subsector's native temporal granularity: entries are filled with zeros where absence of activity is implied or are estimated using Equations 1 and 2. Any remaining gaps that cannot be constrained by these equations are then imputed at the asset level, using backward filling followed by forward filling along the timeseries. After imputation, Equations 1 and 2 are applied to retain mathematical consistency. If data from the same asset cannot be leveraged to fill its timeseries, they are substituted with country-level averages, or, in the absence of a better alternative, global averages.

Once the data are preprocessed, they are re-sampled to monthly estimates, unless they are already provided at that resolution. When raw estimates are at annual resolution, annual emissions are temporally disaggregated according to temporal profiles from EDGAR [*70*], with each Climate TRACE subsector matched with an appropriate EDGAR sector [*71*]. Quarterly estimates are evenly disaggregated to each constituent month. Any other cases are apportioned according to the number of days of overlap between each month and a raw estimate's time span. It should be noted that the only data that are (dis)-aggregated are those which scale with time. For example, the number of animals in agriculture operations does not scale with time, but the emissions per animal do. Finally, asset-level estimates are forward-filled, using the last available month to fill in the same month of subsequent years. Work is ongoing to further refine these methods and incorporate a wider range of data that could bear on imputation and extrapolation.

Supplementary Note 2.0 Climate TRACE-derived emissions estimates

Most emissions estimates included in the Climate TRACE framework are derived from the Coalition itself using sector-specific emissions estimation approaches. This involves the measurement of activities that result in emissions and converting activity to emissions estimates using emission factors. These can broadly be categorized as one of three general approaches: ML and satellite measurements, statistical modeling and satellite measurements, and statistical modeling and reported data. In practice, emissions for a given subsector are usually estimated using a combination of all three estimation approaches. Below, we discuss each type of approach and provide an example of a subsector that utilizes it. Supplementary Dataset 2 summarizes the methodologies used to calculate emissions for each subsector covered by the Climate TRACE dataset, along with links to detailed documentation for each.

Supplementary Note 2.1 Machine learning and satellite imagery approach

Earth observations via satellite remote sensing have emerged as a powerful tool for tracking GHG emissions [*72*], providing high resolution data with broad spatial coverage. Satellites equipped with spectrometers measure atmospheric GHG concentrations by detecting specific light wavelengths, an approach that has been used to measure $CH_4$ emissions from solid waste management [*73*] and $CO_2$ emissions from coal-fired power plants [*74*]. Additionally, satellite imagery can be analyzed to detect land use changes over time, enabling indirect measurement of emissions from deforestation [*75*]. We leverage high spatial and/or temporal-resolution satellite imagery to identify and measure activities that directly contribute to GHG emissions. The ML and satellite imagery approaches follow a bottom-up estimation method whereby satellite imagery is used to visually detect signals that can reliably predict emissions activity. ML models are trained on in situ or reported data to estimate activity with high temporal resolution and global coverage, based on a quantitative relationship between the remotely sensed activity signal and the known activity level. This activity estimation approach is labelled as the “Trained ML model” approach in Column L of Supplementary Dataset 2. This technique enables the detection of detailed temporal fluctuations in activity, resulting in emissions estimates that more comprehensively reflect variability over time. Emission factors, mostly sourced from the Intergovernmental Panel on Climate Change (IPCC), the scientific literature, and industry reports (varies according to subsector; detailed in Supplementary Dataset 2), are applied to activity estimates to derive GHG emissions.

Supplementary Note 2.1.1 Power plant emissions

We use the ML and satellite imagery method to estimate emissions from some major emitting subsectors, including electricity generation (22.2% of global emissions), road transportation (10.81% of global emissions), and iron and steel manufacturing (5.71%). For example, we estimate emissions for combustion power plants that co-release a water vapor plume signal with GHGs, which is detectable with visible satellite imagery [*53*, *63*]. We first assembled a global, geolocated power plant database that includes production metrics like power generation capacity and fuel and prime mover type. To learn the quantitative relationship between the satellite-derived plume signal and electricity generation, ML models (gradient-boosted decision trees and convolutional neural networks; CNNs) are trained on hourly or sub-hourly reported electricity generation data from individual power plants in the USA, Europe, and Australia. The models estimate electricity generation and apply plant-specific emission factors to derive $CO_2$, $SO_2$, nitrogen oxides ($NO_X$), and particulate matter 2.5 (PM2.5) emissions.

This technique can only be applied to plants with wet or mechanical draft cooling towers or wet flue gas desulfurization stacks. The four percent of plants worldwide that do have these structures account for approximately 43% of $CO_2$ emissions from all non-biomass combustion power generation from 2015–2026. To estimate power generation for plants that cannot be detected via satellite signal, country- and fuel-specific average capacity factors are applied to the plant's known capacity, leveraging the harmonized power plant database. These estimates are derived from less spatially specific information and therefore have higher uncertainty, a current limitation of this approach [*53*, *63*].

<u>Supplementary Note 2.1.2 Power plant model validation</u>
Please see Supplementary Dataset 2.

<u>Supplementary Note 2.2 Multiplicative models with satellite measurements</u>
The multiplicative model with satellite measurements approach leverages satellite imagery to identify and characterize emission sources, including factors such as facility type and size. This information is used in multiplicative models to estimate activity levels. Top emitting subsectors estimated with this approach include oil and gas production (6.25% of global emissions), residential onsite fuel usage (5.77% of global emissions), and oil and gas transport (3.34% of global emissions). This estimation approach is labelled as "Multiplicative, Trained ML model" in Column L of Supplementary Dataset 2.

<u>Supplementary Note 2.2.1 Cattle operation emissions</u>
We calculate quarterly facility-level emissions from cattle operations using this type of approach (quarterly emissions are later extrapolated to monthly emissions). Beef and dairy milk production systems are highly emissions-intensive. Cattle produce an estimated 11% of all anthropogenic GHG emissions [*76*, *77*]. Yet, available data on emissions from beef and dairy production are currently coarse (country-level) and are of variable quality, lacking specific attribution to individual facilities. Some regions have permit databases with location data for cattle production, but the majority of cattle operations worldwide remain unidentified.

To enhance emissions transparency within this subsector, we developed a novel global database of cattle operations that unifies disparate country-level reporting and employs ML with remote sensing data to detect previously unknown cattle operations. We use AI models to identify potential cattle operation locations in Sentinel-2 and PlanetScope imagery. Potential locations undergo a human review process to ensure correct identification, which are then fed back into the AI models to improve their performance in identifying other cattle operations. Once individual cattle operations are correctly identified, we estimate the number of cattle per operation using three approaches: (i) some operations report their total head of cattle and we use these numbers as is, such as US Concentrated Animal Feeding Operation (CAFO) permits; (ii) for operations with a known footprint area, we apply a regression model that relates the operation's footprint area to the total head of cattle, using training data from US CAFO permits or studies performed in China; (iii) for operations without a known footprint, we assign a default mean cattle population value specific to the country and cattle type, sourced from the scientific literature and a country's agricultural statistics reports. Operations that are similar to those in the United States or China, where we have known operation footprints, are estimated using the corresponding regression model. Previous studies (e.g., [(*77*, *78*)]) have shown a direct relationship between total head of cattle and total GHG and non-GHG emissions. To derive $CH_4$, $N_2O$, and $CO_2$ emissions, IPCC emission factors are applied to total head of cattle, based on regional characteristics, climate zone, cattle type, and manure management system.

Supplementary Note 2.2.2 Cattle operations model validation
Please see Supplementary Dataset 2.

Supplementary Note 2.3 Multiplicative models with reported data
The multiplicative model with reported data approach approximates a basic bottom-up method that involves applying emission factors to activity data sourced from public and proprietary databases. This estimation approach is categorized as "Multiplicative" in Column L of Supplementary Dataset 2. The highest emitting subsectors estimated with this method include coal mining (2.86% of global emissions), oil and gas refining (1.66% of global emissions), and synthetic fertilizer application (0.93% of global emissions).

Supplementary Note 2.3.1 The aviation sector
The aviation sector is one of the fastest-growing GHG emissions sources [*79*], yet it is exempt from the Paris Agreement. We leverage the International Civil Aviation Organization's (ICAO) bottom-up methodology [*80*] to track flight-level domestic and international aviation emissions, providing greater visibility into this sector. Information on aircraft movement (e.g., origin, destination, aircraft model) sourced from the Official Airline Guides air traffic movement database, a commercially available dataset, is the main dataset used to calculate fuel consumption and GHG emissions. Briefly, the distance between each origin and destination airport is calculated, and the aircraft type for each route is assigned using ICAO data. ICAO fuel consumption tables are used to estimate fuel use by aircraft type. The amount of fuel burned for each flight is calculated and converted to GHG emissions using ICAO's emission factors. For domestic flights, emissions are attributed to the origin country; emissions for international flights are divided equally between the countries involved. Emissions are aggregated by country and airport for the years 2015–2025.

Supplementary Note 2.3.2 Aviation model validation
Please see Supplementary Dataset 2.

Supplementary Note 3.0 Tracking of non-GHGs

Climate TRACE also tracks asset-level emissions of key non-GHGs (carbon monoxide, organic carbon, black carbon, volatile organic compounds, particulate matter, nitrogen oxides, ammonia, and sulfur dioxide) for more than 75% of global emissions, with data available from 2021–2025. This granularity can help identify air pollution hotspots, crucial for understanding exposure patterns that contribute to global mortality, particularly in low- and middle-income countries [*81*].

Two approaches are used to estimate non-GHG emissions. For several subsectors, non-GHG emissions are estimated directly with Equations 1 and 2 of the main manuscript. Emissions factors are derived from the EPA's AP42 [*82*] and the EMEP/EEA Air Pollution Guidebook [*83*]. These sectors include electricity-generation, oil-and-gas-refining, road-transportation, domestic-shipping, international-shipping, residential-onsite-fuel-usage, non-residential-onsite-fuel-usage, petrochemicals-steam-cracking, and cropland-fires.

A country-level, co-pollutant approach is used for estimating air pollution emissions for assets in the remaining sectors. Co-pollutant ratios are of high interest for climatologists as they can provide insight to the anthropogenic processes of a region [*84*]. Here, we utilize them to estimate emissions for the remaining sectors. We estimate a co-pollution ratio between pollutant species and greenhouse gases, defined by:

$$r_{s,g,c,f} = \frac{Em_{s,g,c,f}}{CO_2e_{s,c,f}}$$

where f is the fuel/process responsible for the co-pollution from a sector s, gas g and country c, and $CO_2e$ ($CO_2$-equivalent) for a sector s, gas g, and country c is the global warming potential of all of the $CO_2$, $CH_4$, and $N_2O$ emissions. This ratio is estimated with the Community Emissions Data System (CEDS) [*85*] and Emissions Database for Global Atmospheric Research (EDGAR) [*86*]. It is then multiplied by the $CO_2e$ of Climate TRACE assets to distribute and scale pollutant emissions globally. This process improves the spatial granularity of existing high-quality non-GHG emissions data by leveraging Climate TRACE's source-level location information.

Supplementary Note 4.0 Data Validation Approach

The model performance evaluation process varies by subsector and depends on reference data availability (Supplementary Dataset 2). We aim to evaluate the performance of our predictive models using emissions or activity data from different regions to assess global generalizability, and across different spatial and temporal aggregations to evaluate the agreement with other inventories across scales. For example, the reliability of multiple region-specific linear regression models for predicting total head of cattle are evaluated with several statistical metrics, such as Spearman's rank correlation coefficient, mean squared error (MSE) and the goodness-of-fit measure ($R^2$), RMSE, and mean absolute error (MAE). Additionally, the linear relationship between modeled and reported total head of cattle is evaluated using the same metrics.

Supplementary Note 4.1 Confidence and uncertainty

Like most global GHG inventories, we report probabilistic uncertainty in emissions to quantify inherent uncertainties in our estimates and input data (e.g., measurement errors, activity data variability, and emission factor imprecision), providing users with a quantitative indicator of data reliability. The uncertainty quantification methods vary slightly by subsector; most sectors follow IPCC guidelines [87]. Quantitative uncertainty data are available upon request. Unlike most inventories, we also assign confidence levels to all of our emissions estimates (in addition to input data metrics, including emission factors, activity, and others) to help users understand their reliability and usability for different applications. Unlike most datasets, we also assign qualitative confidence levels (low to high) to our emissions, capacity, and activity estimates, based on the quality and granularity of the data used to derive them. These are published on our interactive emissions map interface. For example, estimates derived from more spatially explicit activity data or emission factors are generally assigned higher confidence levels, while asset-level emissions estimates derived with regional or country-level emission factors would receive lower confidence levels.

## 5.0 Supplementary Results

Climate TRACE global emissions totaled 61.2 billion tonnes of $CO_2$e in 2023 [*89*], a 0.7% increase from 2022 and a 9.2% increase from 2015, when the Paris Agreement was signed. Global $CH_4$ emissions, the second most abundant anthropogenic GHG, reached 391.2 million tonnes in 2023, a 0.2% increase from 2022 and a 9.3% rise from 2015. Subsectors with the highest emissions increases from 2022 to 2023 (100-yr $CO_2$e) include international aviation (27.9%), copper mining (26.4%), domestic aviation (10%), other metals (4.4%), and petrochemical steam cracking (4.1%) [*87*]. This underscores the ongoing challenges with achieving decarbonization, particularly for emissions sectors that are integral to economic stability and infrastructure resilience. Subsectors with the greatest percentage decreases in emissions from 2022 to 2023 (100-yr $CO_2$e) were iron mining (–15.5%), aluminum manufacturing (–7.9%), bauxite mining (–7.1%), chemicals (–6.5%), and international shipping (–3.2%), highlighting areas that could be targeted for further emissions reductions [*89*].

## Supplementary Note 5.1 Methodological details on city-level emissions mitigation analysis

The city-level emissions mitigation analysis was conducted on 500 major urban areas around the world. Using Jenks natural breaks, higher gross domestic product (GDP) cities (30.7% of cities) were defined as GDP per capita ≥ 57,333; Lower-GDP cities (69.3% of cities) were defined as GDP per capita < 57,333.

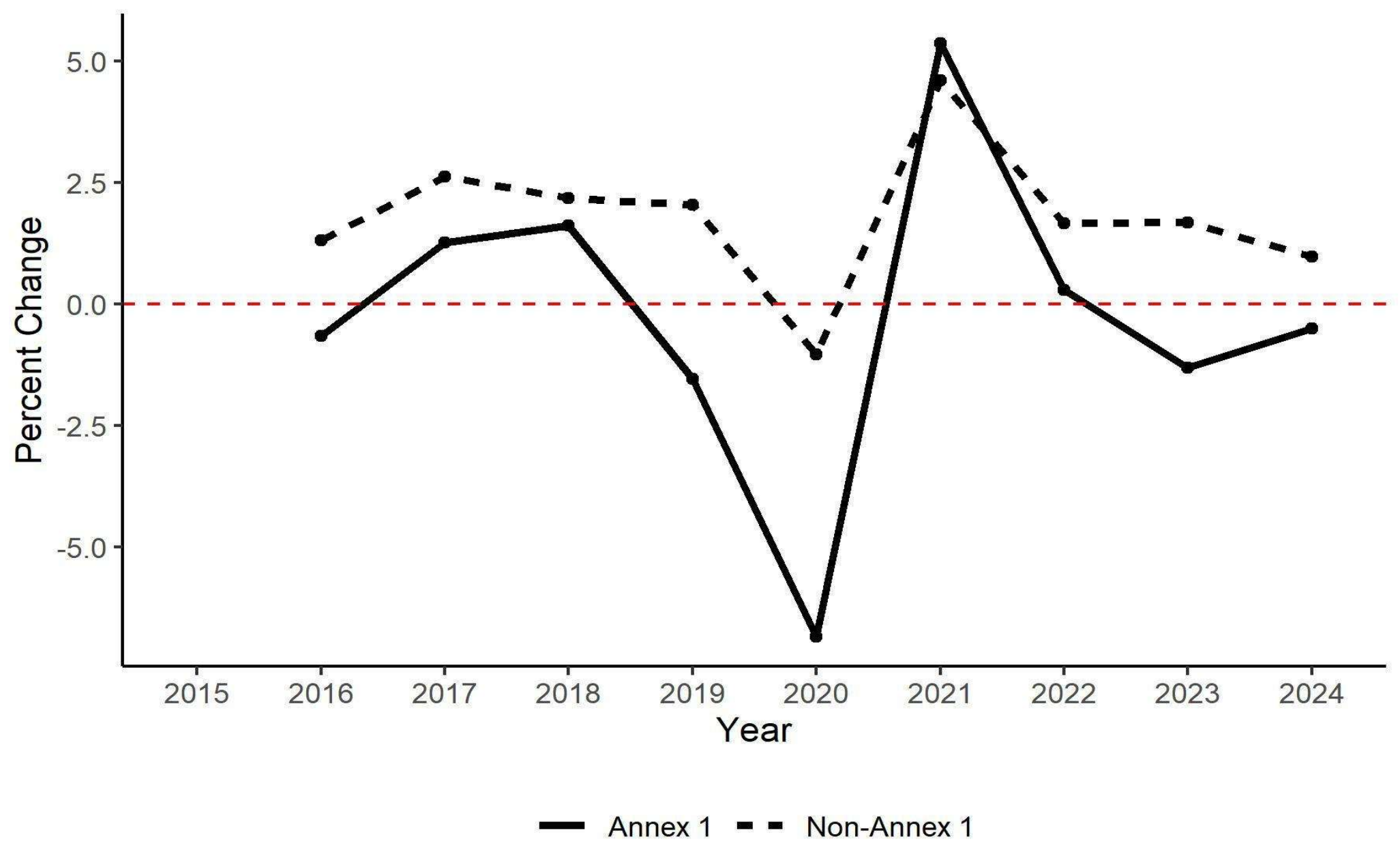

**Supplementary Figure 1**

Climate TRACE emissions data: percent change in annual $CO_2e$ (100-yr) totals for Annex I and Non-Annex I countries between 2015 and 2024.

**Supplementary Table 1.** Acronyms of global greenhouse gas emissions datasets, acronyms definitions, and link to reference(s) used to source information provided in Table 2.

| Dataset acronym | Definition | Reference(s) |
|---|---|---|
| Climate TRACE (v5.4.1) | Climate TRACE (Tracking Real-time Atmospheric Carbon Emissions) | https://climatetrace.org/data |
| Stat. Review of World Energy (2025) | Statistical Review of World Energy | https://www.energyinst.org/statistical-review/resources-and-data-downloads |
| CAMS-GLOB-ANT (v6.2) | Copernicus Atmosphere Monitoring Service GLOB-ANT | https://doi.org/10.5194/essd-16-2261-2024 |
| Carbon Monitor | Carbon Monitor | https://carbonmonitor.org/ |
| CarbonTracker (CT2022) | CarbonTracker | https://gml.noaa.gov/ccgg/carbontracker/ |
| CDIAC-FF | Carbon Dioxide Information Analysis Center - Fossil Fuel | https://rieee.appstate.edu/projects-programs/cdiac/ |
| CEDS v2025.03.18 | Community Emissions Data System | https://zenodo.org/records/15059443, https://zenodo.org/records/15001544 |
| Climate Watch | Climate Watch | https://www.climatewatchdata.org/about/faq/ghg |
| EDGAR (2025) | Emissions Database for Global Atmospheric Research | https://edgar.jrc.ec.europa.eu/dataset_ghg2025 |
| EIA Int'l Energy Outlook (2023) | United States Energy Information Administration International Energy Outlook | https://www.eia.gov/outlooks/ieo/data.php |
| FAOSTAT | Food and Agriculture Organization Statistical Database | https://www.fao.org/faostat/en/#data/GT |
| FFDAS (v2.2) | Fossil Fuel Data Assimilation System | https://ffdas.rc.nau.edu/ |
| GCP-GridFED (v2024.0) | Global Carbon Project – Gridded Fossil Emissions Dataset | https://zenodo.org/records/17467681 |
| GID | Global Infrastructure Emissions Detector | http://gidmodel.org.cn/?page_id=46&lang=en |
| Global Carbon Budget (v2025) | Global Carbon Budget | https://globalcarbonbudget.org/datahub/the-latest-gcb-data-2025/ |
| GRACED | Global Gridded Daily Carbon Dioxide Emissions Dataset | https://carbonmonitor-graced.com/methods.html |
| IEA GHG Emissions from Energy (2024) | International Energy Agency Greenhouse Gas Emissions from Energy | https://www.iea.org/data-and-statistics/data-product/greenhouse-gas-emissions-from-energy |
| ODIAC2024[†] | Open-source Data Inventory for Anthropogenic Carbon Dioxide | https://www.nies.go.jp/doi/10.17595/20170411.001-e.html |
| PIK PRIMAP-hist (v2.7) | Potsdam Institute for Climate Impact Research Potsdam Real-Time Integrated Model for the Probabilistic Assessment of Emission Paths - historical | https://primap.org/primap-hist/ |
| UNFCCC | United Nations Framework Convention on Climate Change | https://di.unfccc.int/time_series |

**Supplementary Table 2**. Comparison of validation approaches and inclusion of self-reported data and uncertainty estimates across global emissions datasets. References are listed in Supplementary Table 1. Datasets with sector-specific emissions estimation methods (e.g., Climate TRACE) can have multiple validation approaches.

| Global Emissions Dataset | Validation approach | Incorporates self-reported data? | Provides uncertainty estimates? | Reference |
|---|---|---|---|---|
| Climate TRACE (v5.4.1) | No validation performed, Component-level validation, In-situ validation, Comparative accuracy evaluation | Yes | Yes | https://github.com/climatetracecoalition/methodology-documents/tree/main. Accessed 22 April 2026. |
| Stat. Review of World Energy (2025) | No validation performed | Yes | No | https://www.energyinst.org/statistical-review/resources-and-data-downloads. Accessed 22 April 2026. |
| CAMS-GLOB-ANT (v6.2) | Comparative accuracy evaluation | Yes | Yes | https://doi.org/10.5194/essd-16-2261-2024 |
| Carbon Monitor | Component-level validation, Comparative accuracy evaluation | Yes | Yes | https://doi.org/10.1038/s41597-020-00708-7 |
| CarbonTracker (CT2022) | Component-level validation | Yes | Yes | https://gml.noaa.gov/ccgg/carbontracker/. Accessed 22 April 2026. |
| CDIAC-FF | Component-level validation, Comparative accuracy evaluation | Yes | Yes | https://doi.org/10.5194/essd-13-1667-2021 |
| CEDS v2025.03.18 | Comparative accuracy evaluation | Yes | Yes | https://essd.copernicus.org/preprints/essd-2020-103/essd-2020-103-manuscript-version4.pdf. Accessed 22 April 2026. |
| Climate Watch | No validation performed | Yes | No | https://www.climatewatchdata.org/about/faq/ghg. Accessed 22 April 2026. |
| EDGAR (2025) | Comparative accuracy evaluation | Yes | Yes | https://doi.org/10.5194/acp-21-5655-2021. Accessed 22 April 2026. |
| EIA International Energy Outlook (2023) | No validation performed | Yes | No | https://www.eia.gov/outlooks/ieo/pdf/IEO2023_narrative.pdf. Accessed 22 April 2026. |
| FAOSTAT | Comparative accuracy evaluation | Yes | No | DOI 10.1088/1748-9326/8/1/015009 |
| FFDAS (v2.2) | Comparative accuracy evaluation | Yes | Yes | doi:10.1029/2009JD013439 |
| GCP-GridFED (v2024.0) | Comparative accuracy evaluation | Yes | Yes | https://doi.org/10.1038/s41597-020-00779-6 |
| GID | Comparative accuracy evaluation | Yes | Yes | http://gidmodel.org.cn/?page_id=2290&lang=en. Accessed 22 April 2026. |
| Global Carbon Budget (v2025) | Component-level validation, Comparative accuracy evaluation | Yes | Yes | https://doi.org/10.5194/essd-17-965-2025 |
| GRACED | Comparative accuracy evaluation | Yes | Yes | https://doi.org/10.1038/s41597-023-01963-0 |
| IEA GHG Emissions from Energy (2025) | Comparative accuracy evaluation | Yes | No | https://www.iea.org/data-and-statistics/data-product/greenhouse-gas-emissions-from-energy#documentation. Accessed 22 April 2026. |
| ODIAC2024 | Comparative accuracy evaluation | Yes | Yes | https://doi.org/10.5194/essd-10-87-2018 |
| PIK PRIMAP-hist (v2.7) | Comparative accuracy evaluation | Yes | No | https://doi.org/10.5194/essd-8-571-2016 |
| UNFCCC | Comparative accuracy evaluation | Yes | Yes; varies by country | https://doi.org/10.1080/17583004.2016.1271256 |

**Table S3**. Share of Climate TRACE emissions estimates that are subject to different accuracy evaluation approaches

| **Climate TRACE sector** | **Share of total global emissions** | **Share of emissions within sector subject to in-situ emissions validation** | **Share of emissions within sector subject to component-level validation** | **Share of emissions within sector subject to comparative evaluation** | **Share of emissions within sector that are not evaluated for accuracy** |
|---|---|---|---|---|---|
| Forestry and Land Use | 21 | 0 | 65 | 34 | 1 |
| Power | 21 | 85 | 0 | 15 | 0 |
| Manufacturing | 14 | 0 | 0 | 58 | 42 |
| Fossil Fuel Operations | 13 | 0 | 0 | 78 | 22 |
| Transportation | 12 | 12 | 0 | 75 | 13 |
| Agriculture | 10 | 10 | 0 | 32 | 58 |
| Buildings | 5 | 0 | 0 | 100 | 0 |
| Waste | 3 | 51 | 0 | 45 | 4 |
| Fluorinated Gases | 2 | 0 | 0 | 0 | 100 |
| Mineral Extraction | <1 | 0 | 0 | 19 | 81 |